\documentclass[10pt,twocolumn,letterpaper]{article}

\usepackage{iccv}
\usepackage{times}
\usepackage{epsfig}
\usepackage{graphicx}
\usepackage{amsmath}
\usepackage{amssymb}
\usepackage{utfsym}
\usepackage{multirow}
\usepackage{booktabs}
\usepackage{graphicx}
\usepackage{subfigure}
\usepackage{authblk}
\definecolor{g}{RGB}{153,204,102}

\usepackage[breaklinks=true,bookmarks=false]{hyperref}

\iccvfinalcopy % *** Uncomment this line for the final submission

\def\iccvPaperID{****} % *** Enter the ICCV Paper ID here

\ificcvfinal\fi

\begin{document}

%%%%%%%%% TITLE
\title{Class-relation Knowledge Distillation for Novel Class Discovery}

\author{Peiyan Gu$^{1,*}$ \quad Chuyu Zhang$^{1,2,}$\thanks{Both authors contributed equally.} \quad Ruijie Xu$^{1}$ \quad Xuming He$^{1,3}$ \\  \vspace{-0.5em}
{\normalsize $^1$ShanghaiTech University, Shanghai, China} \quad
{\normalsize $^2$Lingang Laboratory, Shanghai, China} \quad \\
{\normalsize $^3$Shanghai Engineering Research Center of Intelligent Vision and Imaging, Shanghai, China} \\
{\tt\small \{zhangchy2,gupy,xurj3,hexm\}@shanghaitech.edu.cn}
\vspace{-2em}
}

% \author[1,2]{Chuyu Zhang}
% \author[1]{Peiyan Gu}
% \author[1]{Ruijie Xu}
% \author[1,3]{Xuming He}
% \affil[1]{ShanghaiTech University, Shanghai, China}
% \affil[2]{Lingang Laboratory, Shanghai, China}
% \affil[3]{Shanghai Engineering Research Center of Intelligent Vision and Imaging, Shanghai, China}
% \author{Chuyu Zhang \and Peiyan Gu \and Ruijie Xu \and Xuming He \\
% ShanghaiTech University, Shanghai, China \and Lingang Laboratory, Shanghai, China\\
% % Institution1 address\\
% {\tt\small \{zhangchy2,gpy,xurj3,xuminghe\}@shanghaitech.edu.cn}
% % For a paper whose authors are all at the same institution,
% % omit the following lines up until the closing ``}''.
% % Additional authors and addresses can be added with ``\and'',
% % just like the second author.
% % To save space, use either the email address or home page, not both
% }

\maketitle
% Remove page # from the first page of camera-ready.
\ificcvfinal\thispagestyle{empty}\fi
\vspace{-4em}
%%%%%%%%% ABSTRACT
\begin{abstract}
    We tackle the problem of novel class discovery, which aims to learn novel classes without supervision based on labeled data from known classes. A key challenge lies in transferring the knowledge in the known-class data to the learning of novel classes. Previous methods mainly focus on building a shared representation space for knowledge transfer and often ignore modeling class relations. To address this, we introduce a class relation representation for the novel classes based on the predicted class distribution of a model trained on known classes. Empirically, we find that such class relation becomes less informative during typical discovery training. To prevent such information loss, we propose a novel knowledge distillation framework, which utilizes our class-relation representation to regularize the learning of novel classes. In addition, to enable a flexible knowledge distillation scheme for each data point in novel classes, we develop a learnable weighting function for the regularization, which adaptively promotes knowledge transfer based on the semantic similarity between the novel and known classes. To validate the effectiveness and generalization of our method, we conduct extensive experiments on multiple benchmarks, including CIFAR100, Stanford Cars, CUB, and FGVC-Aircraft datasets. Our results demonstrate that the proposed method outperforms the previous state-of-the-art methods by a significant margin on almost all benchmarks. Code is available at \href{https://github.com/kleinzcy/Cr-KD-NCD}{here}.
 \end{abstract}
\section{Introduction}
The recent development of deep learning has achieved remarkable success in a broad range of visual recognition tasks~\cite{he2016deep,he2017mask,ren2015faster}. However, most traditional methods focus on the closed-world setting, in which all the visual classes are pre-defined. As a result, it is usually difficult to deploy the learned models in realistic settings with potential novel classes. In contrast, human visual systems can efficiently acquire new concepts without supervision based on learned knowledge. %For instance, a human can cluster pandas very well, even if he has never known a panda before. 
Inspired by such an ability, several studies~\cite{han2019learning,hsu2018learning} propose the task of Novel Class Discovery (NCD) which aims to discover novel categories from unlabeled data based on known-class data.
% by unsupervised pretraining on all classes data followed 

\begin{figure}[!tbp]
	\centering
	\includegraphics[width=1.0\linewidth]{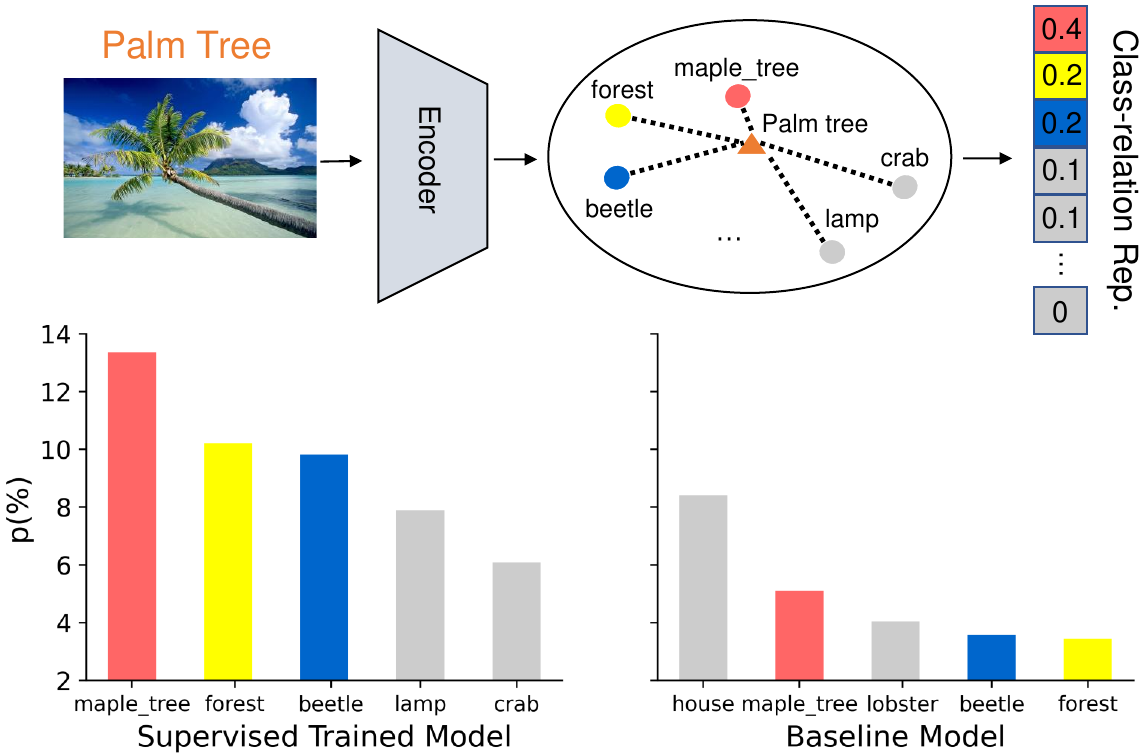}
	\caption{In the upper panel, we apply the encoder and the known class classifier to a novel sample of Palm tree, obtaining a class-relation representation. This representation encodes the relative distances between the representations of novel and known categories. The triangles indicate a novel class, while the circles show known classes. The lower panel shows the averaged class-relation representation for all Palm Tree samples and displays the 5 nearest known classes for both the \textit{known class} Supervised Trained Model and the Baseline Model \cite{fini2021unified}. We observe that the predictions from a trained model on novel class data indicate meaningful class relation (e.g. maple tree, forest), which is lost in the Baseline Model. }
	\label{fig:1}
	\vspace{-1em}
\end{figure}

A key strategy for discovering novel classes is to transfer knowledge in known classes to promote the learning of novel classes. 
%learn a transferable representation, which can better discover novel classes, from known classes.
% semantically meaningful feature representation by transferring knowledge contained in the known classes to the novel ones. 
To achieve this, most existing NCD methods~\cite{han2021autonovel,zhong2021neighborhood,fini2021unified} involve two training stages, including a supervised training stage followed by a discovery training stage. In the supervised training stage, they typically initialize representation by learning from known classes. In the discovery training stage, they transfer the learned knowledge to novel classes via sharing the feature representation space. While they have shown promising results, they are less effective in capturing the relationship between known and novel classes, which limits the scope of shared knowledge and potentially leads to inferior representations of novel classes. Nonetheless, it is difficult to model the semantic relationship between the known and novel classes in the NCD setting as the novel classes are unknown. 

% \footnote{For convenience, we name the first training stage as supervise training stage and the second training stage as the discovery training stage.} 
% For convenience, we name the first training stage as supervise training stage and the second training stage as the discovery training stage.

% 这一段要说明relation的表征形式，然后展示实验观察，呈现当前方法存在的问题。从而启下

% However, our method sKD can effectively reduce the impact of noise and keep meaningful class relations.
% As illustrated in \cite{hinton2015distilling}, the model predictive distribution can represent the class relation. 

To tackle this challenge, we introduce a class-relation representation for a novel class based on its similarity with the known classes. In particular, we leverage a well-known phenomenon of ``dark knowledge"~\cite{hinton2015distilling} and adopt the predicted distribution of a well-trained model to encode the inter-class relationship. To that end, we first train a model on the known classes using supervised learning and then apply the trained model to the data of novel classes. In Fig.~\ref{fig:1} top and bottom-left, we visualize our class-relation representation and the average predictive distribution of a novel class, \textit{palm tree}, respectively. Interestingly, we observe that the distribution often focuses on related or co-occurred classes and hence properly reflects its class relationship. For example, the \textit{palm tree} class is closer to the \textit{maple} and  \textit{forest} classes. However, a typical NCD baseline, which fine-tunes the pre-trained model, is unable to maintain such a similarity structure, as indicated by the example shown in Fig.~\ref{fig:1} bottom-right. Here the \textit{palm tree} class is more similar to the \textit{house} class, which is less reasonable.  

%Notably, upon comparing the two predictive distributions, we discover that the class relation represented by the supervised trained model on known classes is more reasonable than the baseline model, which performs novel class discovery by fine-tuning the supervised trained model. For example, on the baseline, the palm tree class is closer to the house class and further away from the maple tree and forest classes, while on the supervised model,  We speculate that during the discovery training stage, the unsupervised learning of novel classes may influence the representation and classifier, leading to undesirable class relations for some novel classes.

% the classifier activation of a novel class on a model that has been pre-trained on known classes can represent the class relation. 
% Unfortunately, due to the noisy clustering of novel classes, the class relation is lost during joint training on known and novel class data. 

% 引用一些class relation的paper

Motivated by the above observation, we propose a novel class-relation knowledge distillation framework for the task of novel class discovery. Our framework utilizes the class relation represented by the supervised trained model to regularize the learning of novel classes in the discovery training stage, thus preserving the meaningful class relation knowledge and promoting knowledge transfer. Moreover, to provide a flexible knowledge transfer scheme for each data sample, we develop a simple but effective learnable weight function for the regularization, which allows us to adaptively transfer knowledge based on the similarity between a novel class sample and known classes.
%assigns more/less weight to semantically similar/dissimilar novel samples, respectively.

Specifically, we instantiate our framework with a two-head network architecture that includes an encoder and two classifier heads for the known and novel classes, respectively. We first initialize the feature representation through supervised learning on the known classes and then discover novel classes by minimizing a hybrid learning loss. Our loss consists of three terms: 1) a standard cross-entropy loss on the labeled data, which extracts semantic knowledge from the known classes; 2) an unsupervised clustering loss on the novel class data; and 3) a weighted Kullback–Leibler (KL) regularization term for distilling the class relation knowledge from the supervised trained model into the discovery of novel classes. Here the strength of each KL regularization term is controlled by a weight measuring the similarity between a novel sample and the known classes, which is derived from the predicted distribution on the known classes by the model in the discovery training stage.

% mutual information between the labeled and unlabeled data, aiming to facilitate transferring knowledge on classifiers from the known classes to the novel. 
% To estimate the mutual information, we develop an approximate sampling method along with a projection operator for the OWSSL task. Given this loss, we train our classifier network for both known and novel classes in an end-to-end manner. 

To validate the effectiveness of our method, we conduct extensive experiments on four datasets, including CIFAR100, Stanford Cars, CUB, and FGVC-Aircraft. The results show that our performances surpass the previous state of the art by a large margin in most cases, demonstrating the efficacy of our novel design of learning framework. 
In summary, our main contributions are three-fold:
 \begin{itemize}
 	\setlength{\itemsep}{0mm}
     \item  We propose a simple and effective learning framework to facilitate knowledge transfer from the known to novel classes, which provides a new perspective to tackle novel class discovery problems. 
     \item We propose a novel regularization strategy to capture class relation between known and novel classes via the classifier output space, and develop a simple but effective learnable weight function to adaptively transfer knowledge based on the strength of class relation.
     \item Our method significantly outperforms previous works on various public benchmarks, illustrating the efficacy of our design.
 \end{itemize}

\section{Related Work}
% 从四个或者三个方面来介绍related work: 1. novel class discovery这块是重点，说明发展脉络，优缺点。最后指出他们的缺点，然后引出我们的方法。 2. semi-supervised learning： 这一块主要是阐述区别，novel class discovery和他的区别是什么，我们的问题存在的必要性。这一块，我觉得可要可不要，历史的paper大都牵涉这一块，前人有解释setting的必要，如今可能不需要。但是general novel class discovery setting是在semi-supervised learning的框架下提出的，这可能是需要介绍semi-supervised learning的唯一原因。3.Transfer learning： 因为我的motivation也是transfer knowledge，所以需要阐述下这一块工作，并说明我们方法的区别是什么。4. Mutual information maximization for representation learning: 这个和我的核心贡献相关，要说明我们的不同之处是什么。不是简单搬运。
\begin{figure*}[!tbp]
\centering
\includegraphics[width=0.75\linewidth]{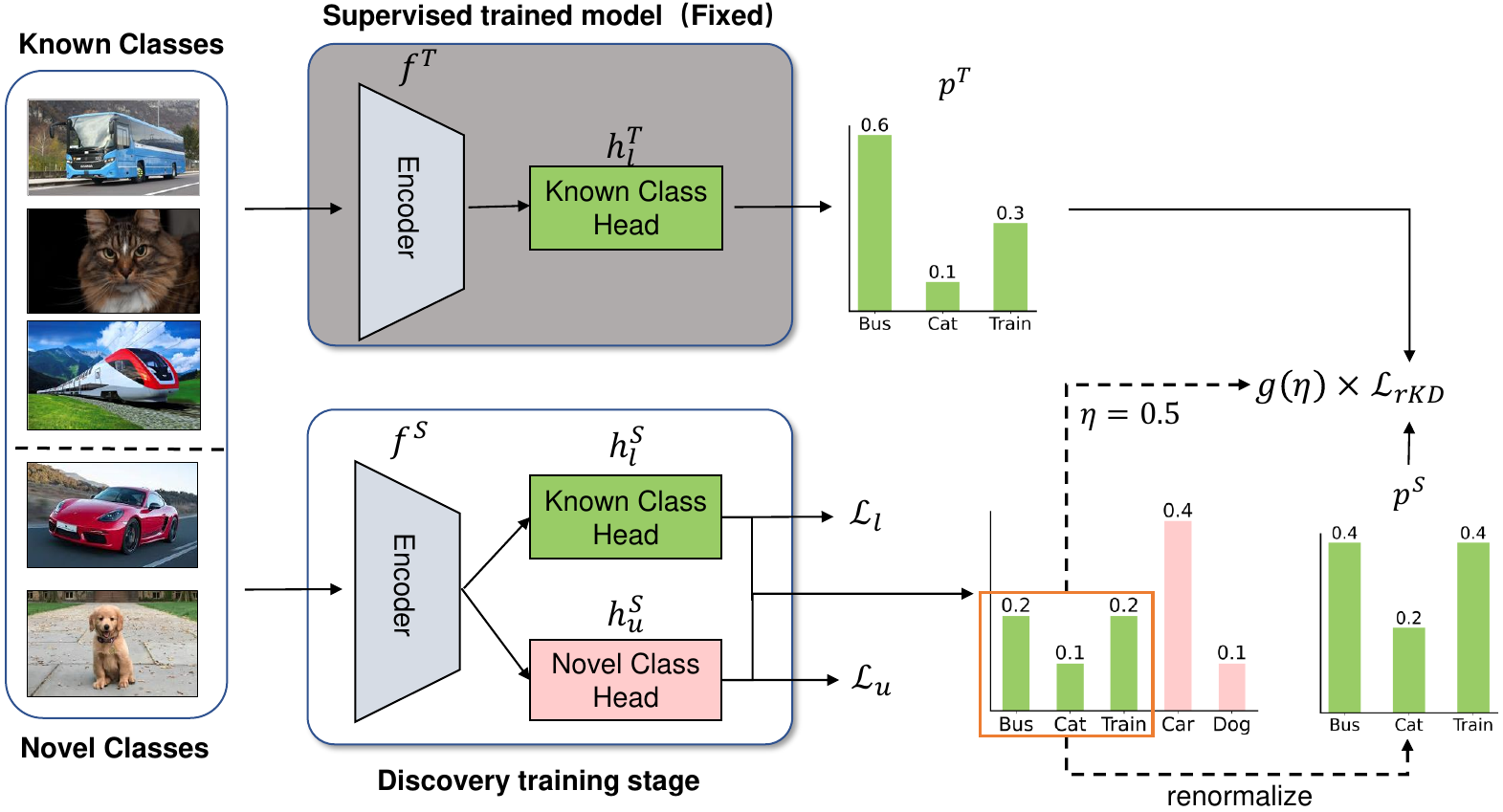}
\caption{The overview of our class relation knowledge distillation framework. We first train a model by supervised learning on the known classes. Then we discover novel classes by jointly learning known and novel classes. To maintain meaningful relation information, we utilize the class relation represented by the supervised trained model to adaptively regularize the learning of novel classes in the discovery stage. $\eta$ represents the semantic similarity between the novel class sample and known classes, and $g(\eta)$ is the learnable weight function. And the supervised trained model is omitted in the inference.}
\label{fig:method_overview}
\vspace{-1em}
\end{figure*}

\paragraph*{Novel class discovery:} The idea of novel class discovery was initially explored in~\cite{hsu2018learning, hsu2018multi}, which performs transfer learning across domains and tasks, and utilizes predictive pairwise similarity as the knowledge for clustering. The standard NCD problem was formalized by~\cite{han2019learning}, aiming to cluster novel classes with the help of known classes.  
Most NCD methods attempt to transfer knowledge from known to novel classes by learning a shared representation, and can be categorized into two groups according to their clustering methods. 
The first group \cite{hsu2018learning,han2019learning,zhong2021neighborhood,zhong2021openmix,zhao2021novel} typically explores pair-wise similarity for clustering. For example, KCL~\cite{hsu2018learning} learns a pair-wise similarity network to predict the similarity of two instances. RankStats~\cite{han2021autonovel} propose robust rank statistics to measure the similarity of two data in their representation space. Furthermore, DRNCD~\cite{zhao2021novel} presents dual rank statistics that focus on local part-level information and overall characteristics. In addition to pair-wise similarity objective, NCL~\cite{zhong2021neighborhood} and Openmix~\cite{zhong2021openmix} utilize contrastive learning and mixup strategy to promote the representation learning of novel classes. 
The second group~\cite{han2019learning,zhong2021neighborhood,yang2022divide} introduces self-labeling for clustering novel classes. Especially, DTC~\cite{han2019learning} utilizes deep embedding clustering to discover novel classes. UNO~\cite{fini2021unified} adopts the Sinkhorn-Knopp algorithm to generate pseudo labels. Unlike above, ~\cite{chi2021meta} solves novel class discovery from a meta-learning perspective. Recently, several works\cite{cao2021open,vaze2022generalized} have expanded the conventional novel class discovery problem to more practical scenarios where unlabeled data sets consist of known and novel classes.
%A few recent methods consider transferring knowledge through a joint label space~\cite{chi2021meta,zhao2021novel}.
% And recently, ~\cite{chi2021meta} solve novel class discovery from a meta-learning perspective.
% concatenates the logits of labeled and unlabeled data
% Although these methods have achieved significant success, they are limited to transferring knowledge due to the classifier learning being disjoint.
% Furthermore, ~\cite{zhao2021novel} present dual rank statistics that focus on local part-level information and overall characteristics.
% NCL~\cite{zhong2021neighborhood} applies contrastive learning to learn a discriminative representation, and Openmix~\cite{zhong2021openmix} utilizes the mixup strategy to combine the known and novel classes. 
% ComEx~\cite{yang2022divide} proposes a divide-and-conquer strategy to recognize both known and novel classes. 
% and apply a unified classification loss for both labeled and unlabeled data

Although these methods have achieved some success, few of them consider the potential relationship between the known and novel classes in the label space during model learning. In this paper, we model the relations between known and novel classes based on model predictions, and propose a novel knowledge distillation method to transfer class-relation knowledge, thus improving the representation learning of novel classes.

%Most of them transfer knowledge by sharing representation space and focus on obtaining better pseudo labels instead of better-transferring knowledge between seen and unseen classes.

% ~\cite{han2021autonovel,zhao2021novel} propose robust rank statistics to measure the similarity of two data on representation space. After that, ~\cite{zhong2021neighborhood,zhong2021openmix} applied contrastive learning and mixup to learn better representation for unseen classes. ~\cite{fini2021unified} utilizes Sinkhorn-Knopp algorithm to generate pseudo labels, and applies a unified classification loss for both labeled and unlabeled data. ~\cite{chi2021meta} solve novel class discovery from a meta-learning perspective. Most of them~\cite{han2021autonovel,zhong2021neighborhood,fini2021unified,yang2022divide} transfer knowledge by sharing representation space, and focus on obtaining better pseudo labels instead of better-transferring knowledge between seen and unseen classes.

% Apart from transferring knowledge by sharing representation space, some methods transfer knowledge through a joint label space. Specifically,
% And recently, ~\cite{chi2021meta} solve novel class discovery from a meta-learning perspective.
% concatenates the logits of labeled and unlabeled data
% Although these methods have achieved significant success, they are limited to transferring knowledge due to the classifier learning being disjoint.
% Furthermore, ~\cite{zhao2021novel} present dual rank statistics that focus on local part-level information and overall characteristics.

% 阐述一下transfer learning主要是做什么，他们与我的方法区别是什么。
% 
\vspace{-1em}
\paragraph*{Knowledge distillation:} 
% Knowledge distillation~\cite{hinton2015distilling,gou2021knowledge,park2019relational,wang2021knowledge} aims to transfer knowledge from a teacher model to a student model. Usually, they can be categorized into two groups based on the knowledge transferred in model prediction space or representation space. For methods transferring knowledge in model prediction space, they assume that the probability distribution produced by the teacher model provides more information about which classes are more similar to the predicted class than the one-hot ground truth. Therefore, they utilize the semantic meaningful probability distribution produced by the teacher model to supervise the learning of the student model. For another group, they \cite{ahn2019variational,ian2019contrastive} argue the representation learned by the teacher network contains rich structural information, and they propose to distill knowledge from teacher to student in the representation space by maximizing mutual information.

Knowledge distillation~\cite{hinton2015distilling,gou2021knowledge,park2019relational,wang2021knowledge} aims to transfer knowledge from a teacher model to a student model. Typically, it can be categorized into two groups based on whether knowledge is transferred in the model prediction space or the representation space. In the first group~\cite{hinton2015distilling,zhao2022decoupled}, the methods often assume that the probability distribution produced by the teacher model provides more information about which classes are more similar to the predicted class than the one-hot ground truth. Therefore, they use the semantically meaningful probability distribution produced by the teacher model to supervise the learning of the student model. The second group of methods~\cite{ahn2019variational,tian2019contrastive} instead argue that the representation learned by the teacher network contains rich structural information. These methods propose to distill knowledge from the teacher to the student in the representation space by maximizing mutual information. We refer the readers to ~\cite{wang2021knowledge} for a more comprehensive survey.

In contrast to the aforementioned works, where the teacher and student networks share the same category space, our approach involves knowledge distillation between known and novel classes, which are in separate category spaces. Specifically, we distill knowledge from a model trained on known classes to a model trained on both known and novel classes, with the goal of transferring knowledge from known classes to novel classes.
% Hinton et al.~\cite{hinton2015distilling} named it as dark knowledge.

% Transferring knowledge between classes has been explored in many different learning paradigms. For instance, transfer learning~\cite{zamir2018taskonomy,zhuang2020comprehensive,weiss2016survey} typically aims to transfer knowledge from the source domain to the target domain. 
% Among them, the idea of~\cite{ahn2019variational} is most relevant to our method, in which they transfer knowledge by retaining high mutual information between the layers of the teacher and student networks.

% \textcolor{red}{Inspired by ~\cite{ji2019invariant} but differ from them, which aims to learn an invariant representation and maximize the mutual information between the class assignments of each data pair, we aim to maximize the mutual information between labeled and unlabeled data and transfer knowledge between classifier.}

\section{Method}

%xhe开始：加一两句要解决什么问题
As shown in Fig.\ref{fig:method_overview}, our model training is divided into two stages. In the supervised training stage, we train our model with known class data to obtain an initial feature representation, which contains meaningful semantic information, thus providing a good initialization for clustering novel classes. In the discovery training stage, we train the model with both known and novel class data, and adopt the typical cross-entropy loss and self-labeling loss to learn the known and novel classes, respectively. In addition, to better transfer knowledge, we propose a novel adaptive regularization term to distill relation knowledge from the known classes pretrained model. We will discuss the learning of our framework in detail in the following.

% and expect to transfer knowledge between known and novel classes by sharing the feature representation space. However, due to the lack of using the semantic relation between known and novel classes, such a way is ineffective to transfer knowledge. Empirically, as shown in Figure 1, the model of the discovery stage drops some class relation information compared to the first supervise trained model. We assume that the class relation represented by the supervised pretrained model is important and can help to cluster novel classes. Consequently, to improve knowledge transfer, we propose a novel adaptive regularization term to distill knowledge from the initialized model to the discovery training stage. 

In this section, we first introduce our novel class relation distillation framework in Sec.~\ref{sec:framework}. Then we describe the losses on the labeled data for the known classes and unlabeled data for novel classes in Sec.~\ref{sec:relationlabeing}. Finally, we present our novel relation knowledge distillation loss in Sec.~\ref{sec:kd}, which is the core design of our method.

\subsection{Class Relation Distillation Framework}\label{sec:framework}
To introduce our framework, we first present the problem setting of NCD and notations.
% \subsection{Problem Setup}\label{sec:setup}
%xhe开始：加一下和NCD的标准设定的关系。
The training dataset consists of two parts: a labeled known classes set $\mathcal{D}^l = \{x^l_i, y^l_i\}_{i=0}^{|\mathcal{D}^l|}$ and an unlabeled novel classes set $\mathcal{D}^u = \{x^u_j\}_{j=0}^{|\mathcal{D}^u|}$. Here $x, y$ represent the input data and the corresponding label, respectively. We use $Y^l=\{1, 2, ...,C^l\}$ and $Y^u=\{C^l+ 1, C^l+2, ...,C^l+C^u\}$ to represent the category space of known and novel classes, respectively.

% and $Y^u=\{C^l+ 1, C^l+2, ...,C^l+C^u\}$ to represent the category space of novel classes.
%, i.e. $Y^l\cap Y^u=\emptyset$. 
%The goal is to cluster $\mathcal{D}^u$ with the presence of $\mathcal{D}^l$. 
%xhe这个目标定义的不清楚，聚类应该只是手段，要回到class discovery上面。

% \subsection{Method Overview}\label{method:overview}

%xhe这部分更像一个方法总体框架的介绍而不是overview。Overview需要简洁和突出设计思想。

% our model consists of an encoder and two heads, corresponding to the classification head for labeled data and the clustering head for unlabeled data. 
% 我们的模型训练分成两步，第一步，我们用已知类数据初始化我们的模型，得到初始的特征表示。这样的特征表示包含一些语义信息，为新类的聚类提供不错的初始化。第二步，我们用已知类和未知类数据训练模型，通过共享特征空间，来实现未知类和新类之间的知识迁移。如图1所示，第二阶段的联合训练相对于第一阶段，丢掉了一些class relation信息。我们认为这些class relation比较重要，能够帮助novel class学习，因此，为了提升知识迁移，我们提出一个novel regularization term，将初始化模型中的知识蒸馏到第二阶段中，我们在后续详细讨论这一部分。具体而言，

We adopt a common model architecture for NCD, consisting of an encoder, denoted by $f$, along with two cosine classifier heads: $h_l$ for known classes and $h_u$ for novel classes. The encoder can be a standard convolutional network (CNN) or Vision Transformer (ViT) \cite{dosovitskiy2020image}.
Given an input image from a known or novel class, we first project it into an embedding space through the shared encoder. Then we normalize the embedding and feed it to the known and novel class head. Note that no matter whether an input is known or novel, it will go through two heads to generate two outputs. Finally, we concatenate the two outputs as the final prediction.
% In addition, to utilize the knowledge learned by the supervised trained model, we fix the supervised pretrained model and forward it. 
The forward process can be written as:
\begin{equation}\label{pyx}
    p(y|x) = \text{Softmax} ((h^S_l(f^S(x))\oplus h_u^S(f^S(x))) / \tau)
\end{equation}
where superscript $S$ denotes the model in discovery training stage, $p(y|x) \in \mathbb{R}^{C^l + C^u}$ is the model predictive distribution, and $\tau$ is the temperature of the softmax function. 

%in the inference, only the discovery trained model is preserved.
To discover novel classes, we begin by initialize our representation ability using supervised learning with known classes, then discover novel class by training on known and novel class data jointly. In the discovery training stage, our objective function consists of three terms: 1) a supervised loss for known class data, 2) an unsupervised loss for novel class data, and 3) a class-relation Knowledge Distillation loss for novel class data. The overall loss can be written as:
\begin{equation}
    \mathcal{L} =  \mathcal{L}_{l} + \alpha \mathcal{L}_{u} + \beta \mathcal{L}_{rKD}
\end{equation}
where $\mathcal{L}_{l}$ is the standard supervised loss on known classes data, $\mathcal{L}_{u}$ is the unsupervised clustering loss for novel classes data, and $\mathcal{L}_{rKD}$ is our relation Knowledge Distillation loss. Here $\alpha, \beta$ are the weighting factors.%, and $\alpha$ is set to 1.

\subsection{Loss for known and novel classes}\label{sec:relationlabeing}
We now present the first two loss terms for the known and novel class data in the discovery training stage, respectively. 
% We note that the choice of those two losses is orthogonal to our method and here we summarize some widely-used loss functions. 
% For the supervised loss on the labeled data, we adopt the standard cross-entropy loss. For the unsupervised loss on the unlabeled data, we evaluate our method with two options: 1) a pairwise similarity loss~\cite{hsu2018learning, han2021autonovel,cao2021open} %, which minimizes the distance of a pair of similar data; or 
% and 2) a relation-labeling loss~\cite{asano2020relation, fini2021unified}.
For the supervised loss on the known classes, we adopt the standard cross-entropy loss. For the unsupervised clustering loss on the novel classes data, we adopt the widely used self-labeling loss~\cite{asano2020self, fini2021unified}, which assigns pseudo labels for novel classes data by solving an optimal transport (OT) problem, then utilizes the generated pseudo label to self-train the model.
% \paragraph*{Pairwise similarity loss~\cite{chang2017deep, hsu2018learning}:} 
% The pairwise similarity loss encourages grouping a pair of similar data, thus learning compact representation for unlabeled data. Specifically, given a batch of $B$ unlabeled data, we compute the embedding $z^u=f(x^u)$ and prediction $ \mathbf{y}^u = p(y^u| x^u) $. For each unlabeled data, %to get its pairwise pseudo label, 
% we find its nearest neighbor in the embedding space from the $ B $ unlabeled data and denote the nearest neighbor of $z^u_i$ as $\hat{z}^u_i$. The pairwise loss~\cite{cao2021open} can be written as:
% \begin{equation}
%     \mathcal{L}_{u} = \frac{1}{|\mathcal{D}^u|}\sum_{i=0}^{|\mathcal{D}^u|} -\log (\mathbf{y}^u_i)^T \mathbf{\hat{y}}^u_i
% \end{equation}
% To prevent all the unseen classes from collapsing to a single cluster,~\cite{cao2021open} also introduces a simple entropy regularization term to regularize the size of the cluster.
% \vspace{-0.7em}

% \paragraph*{relation-labeling loss~\cite{asano2020relation}:}
% The relation-labeling loss first generates a pseudo label for unlabeled data, then utilizes the generated pseudo label to relation-train the model. 
Specifically, such a self-labeling process assumes that the data of novel classes are equally partitioned into clusters and utilizes Sinkhorn-knopp algorithm to find an approximate assignment. We denote $\mathbf{y}^q=q(y^u|x^u)$ as pseudo label, $\mathbf{y}^p=p(y^u|x^u)$ as model's prediction, and $\mathbf{y}^p, \mathbf{y}^q \in \mathbb{R}^{C^u \times 1}$. Let $\mathbf{Q}=[\mathbf{y}^q_1, \mathbf{y}^q_2,,,\mathbf{y}^q_B] \frac{1}{B}$, $ \mathbf{P}=[\mathbf{y}^p_1, \mathbf{y}^p_2,,,\mathbf{y}^p_B] \frac{1}{B}$ be the joint distribution of $B$ sampled data. We estimate $\mathbf{Q}$ by solving an OT problem:
\begin{align*}
\langle\mathbf{Q},-&\log\mathbf{P}\rangle_F \\
%\mathbf{U}(\bm{\mu}, \bm{\nu}) &= 
\text{s.t.}\enspace \mathbf{Q}\in \{\mathbf{Q}\in \mathbb{R}^{C^u\times B}_+ |  \mathbf{Q}\mathbf{1}_{B} = &\frac{1}{C^u}\mathbf{1}_{C^u}, \mathbf{Q}^\top \mathbf{1}_{C^u}=\frac{1}{B}\mathbf{1}_B \} \label{eq:relax_ot_w_2}
\end{align*}
where $\langle,\rangle_F$ is the Frobenius inner product. 
We refer readers to~\cite{cuturi2013sinkhorn,asano2020self} for the details of optimization. The optimal $\mathbf{Q}$ is the pseudo label of unlabeled data and we denote the optimal pseudo label as $q^\ast (y^u|x^u)$. The self-labeling loss is:
\begin{equation}
    \mathcal{L}_{u} = \frac{1}{B}\sum_{i=1}^{B}-q^\ast (y_i^u|x_i^u)\log p(y_i^u|x_i^u)
\end{equation}

%The supervised loss on the known classes ensures the encoder extracts semantic meaningful representation. 
To transfer knowledge between known and novel classes, previous methods~\cite{han2019learning,zhao2021novel,cao2021open} couple the learning of known and novel classes by sharing the encoder $f^S$. The models are typically learned by optimizing the supervised cross-entropy loss on labeled data and the self-labeling loss on unlabeled data. 
This parameter sharing and the jointly-optimized model allow them to learn representations helpful for novel class clustering. However, such an implicit knowledge transfer method is incapable of fully utilizing the knowledge contained in known classes for the clustering of novel classes. Below, we introduce a novel class relation knowledge distillation term to constrain the model learning in the discovery training phase, resulting in a better representation and learning of the novel classes.

%Moreover, as shown in Fig.\ref{fig:1}, the learned meaningful class relations represented by the supervised trained model are lost in the joint representation space due to learning of novel classes. 

% due to the noise contained in the pseudo label for unlabeled data, those methods tend to be biased toward the seen classes, leading to an inferior representation for the unseen classes. What's more, the classifier learning of seen and unseen classes is disjoint, neglecting the potential relation constraint via the model output space. Below, we introduce a regularization term to constrain the model learning and facilitate the representation learning of the unseen classes.

\subsection{Class-relation Knowledge Distillation}\label{sec:kd}
For more effective knowledge transfer, we introduce a class-relation representation based on the output distribution of a model on the known classes classifier. Such a distribution encodes the similarity structure between a novel class data and the known classes. However, current methods transfer knowledge by sharing an encoder, which is less effective in capturing class relations for novel classes. In particular, as shown in Fig.~\ref{fig:1}, we find that there is a meaningful class relation contained in the supervised pretrained model, but the class relation is less meaningful for the baseline model. We speculate that such a class relation is important for learning a good representation of novel classes, while the conventional discovery training stage inadvertently changes the representation space and weakens the relations between known and novel classes.

% not concat?
Therefore, we propose a novel relation Knowledge Distillation (rKD) loss to distill the knowledge contained in the supervised trained model to enhance the learning of novel classes. Our rKD loss regularizes the model learning process in the discovery stage, preventing the model from losing meaningful class relations. %, which are vitally important for clustering novel classes. 
Moreover, since the class relation may vary for different novel class samples, we consider an adaptive regularization scheme based on the strength of the class relation. For instance, for novel class samples that are more similar to known classes, we impose a larger regularization weight. To achieve this, we propose a simple but effective learnable weight function to control the regularization effect for different novel samples. In the following section, we formally introduce our novel rKD loss and weight function.

\paragraph*{Knowledge Distillation:} In the discovery training stage, for novel classes, we encourage the model to learn discriminative representation while maintaining the relations to the known classes. To this end, we keep the supervised trained model and feed each novel-class sample into the model to compute an initial relation representation. Similarly, we use the model $f^S$ to obtain the current class relation. This process can be written as follows:
\begin{align}
    \vspace{-0.5em}
    p^T &= \text{softmax}( (h^T_l(f^T(x^u))) / t) \in \mathbb{R}^{C^l} \\
    p^S &= \text{softmax}( (h^S_l(f^S(x^u))) / t) \in \mathbb{R}^{C^l} 
    \vspace{-0.5em}
\end{align}
where $p^T,p^S$ denotes the relation representation of the supervised and discovery-stage model respectively. $t$ is the temperature used to soften relation representation. A higher value in $p^T$ means the novel class sample is semantically closer to the corresponding known class. To regularize the model learning in the discovery training stage, we propose a novel knowledge distillation (KD) loss term based on the KL divergence between $p^S$ and $p^T$:
% our novel relation knowledge distillation regularization term for novel class samples:
\begin{equation}
    \mathcal{L}^u_{rKD} = \frac{1}{B}\sum_{i=1}^{B}\text{KL}(p_i^T||p_i^S)
    \vspace{-0.5em}
\end{equation}
%where $B_u$ is the number of novel samples in a batch. 
$p^T$ is fixed in the discovery training stage. The loss $\mathcal{L}^u_{rKD}$ regularizes the learned representations to maintain relative relation between novel class samples and known classes represented by the supervised trained model. 
%What's more, we also can regularize the learning of known classes by relation knowledge distillation.
\vspace{-1em}
\paragraph*{Learnable weight function:} Typically, novel-class samples have varying semantic similarities and for the semantic dissimilar samples, the relation represented by $p^T$ is relatively noisy. This requires an adaptive regularization for the novel-class data. To tackle this, we propose a simple but effective learnable weight function to control the regularization strength for different novel class samples. For a novel class sample, we first utilize the sum of known classes' probability in Eq.\ref{pyx} to represent the relation strength to the known classes. Formally, the relation strength for the ith sample in a batch is:
\vspace{-0.5em}

\begin{equation}
    \eta_i = \sum_{k=1}^{C^l}p_{i,k}(y^u_i|x^u_i)
\end{equation}
where $p_{i,k}(y_i^u|x^u_i)$ denotes the sample $x^u_i$ probability on class $k$.
Higher $\eta$ indicates a stronger semantic relation with the known classes. Then, we develop a learnable weight function $g$ as a positive correlation function about $\eta$. In this work, we adopt a simple design that computes a normalized relation strength over the batch:
\begin{equation}
    g(\eta_i) = \text{Norm}(\eta_i) = B\frac{\eta_i}{\sum_{j=1}^{B}\eta_j}
\end{equation}
In our experiments, the batch size is large enough to ensure that the statistics of the mean are stable. With our learnable weight function, the adaptive relation knowledge distillation loss for novel classes can be written as:
\begin{equation}\label{L_rkd}
    \mathcal{L}_{rKD} = \frac{1}{B}\sum_{i=1}^{B}g(\eta _i)\text{KL}(p_i^T||p_i^S)
\end{equation}
With our novel adaptive class relation knowledge distillation regularization term, our model can cluster novel classes and maintain the semantic meaningful representation structure simultaneously.
% As our weight function $g(\eta)$ relies on the model's output, it is learnable. When minimizing Equ.\ref{L_rkd}, we simultaneously decrease the weight function's value and the KL constraint term. Thanks to normalization, the weight function cannot converge to the trivial solution of all zeroes. Essentially, for samples with a larger KL divergence, we optimize relations by decreasing the KL term and the weight function simultaneously. This procedure downweights learning on difficult examples. Given the weight function is the sum of known class probabilities, decreasing the weight function means moving this sample away from the known classes. Conversely, for samples with a smaller KL divergence, we relatively increase the weight function, i.e., pulling this sample closer to the known classes. Consequently, this empowers the model to learn shared semantic information between the novel and known classes more effectively.

The utilization of this proposed design confers three advantages. Firstly, the function $g(\eta)$ exhibits a positive correlation with $\eta$, thereby directing the model to prioritize the KL loss of samples that share a higher similarity with the known classes. Secondly,  given that the mean value of $\eta$ may vary across datasets, the normalization procedure ensures that the weight values $g(\eta)$ remain uniform across datasets, leading to a more consistent application of the hyperparameter of $\mathcal{L}_{rKD}$. Finally, the learnability of our weight function allows the weight function to adapt dynamically based on the relation's learning dynamics. In particular, for samples with a higher KL divergence, we optimize relations by simultaneously decreasing the KL divergence term and weight function to downweight the learning on challenging examples. Conversely, for samples with a lower KL divergence, we relatively increase the weight function, empowering the model to learn shared semantic information between the novel and known classes more effectively. We ablate the design of the weight function in the experiments.

% 相比于一般的权重函数，我们的权重函数有两个特点归一化使得

% Moreover, the normalization process ensures that our weight function does not converge to a trivial solution of all zeros. Intuitively, when dealing with a new class sample whose relationship is challenging to learn, we believe that its relationship representation is noisy. Therefore, we decrease its weight to push the novel sample away from the known classes. Conversely, we increase the weight of samples that are easy to learn to bring the novel class sample closer to the known class. This way, the model can transfer knowledge more effectively to novel samples. We ablate the design of the weight function in the experiments.

% In conclusion, our novel adaptive relation knowledge distillation loss is:
% \begin{equation}
%     \mathcal{L}_{rKD} = \mathcal{L}_{rKD}^l  + \sigma \mathcal{\hat{L}}_{rKD}^u
% \end{equation}
% $\sigma$ is the balance factor. 

% \subsection{Model Training}
\section{Experiments}
\begin{table}[!t]
    \centering
    \caption{The details of dataset split.}
    \resizebox{0.4\textwidth}{!}{
    \begin{tabular}{ccccc}
    \toprule
    \multirow{2}{*}{Dataset} & \multicolumn{2}{c}{Known} & \multicolumn{2}{c}{Novel}              \\ \cline{2-5} 
                             & Images          & Classes   & Images       & \multicolumn{1}{c}{Classes} \\ \midrule
    CIFAR100-20              & 40.0k           & 80        & 10.0k        & 20                          \\
    CIFAR100-50              & 25.0k           & 50        & 25.0k        & 50                          \\
    Stanford Cars              & $\approx$4.0k           & 98        & $\approx$4.1k        & 98                          \\
    CUB           & $\approx$3.0k    & 100        & $\approx$3.0k & 100                          \\ 
    FGVC-Aircraft              & $\approx$3.3k           & 50        & $\approx$3.3k        & 50                          \\ \bottomrule
    \end{tabular}
    }
    \label{exp:dataset}
\vspace{-1em}
\end{table}

\begin{table*}[!t]\small
\caption{Comparison with the SOTA methods on the unlabeled training set of the CIFAR100, Stanford Cars, CUB, and Aircraft datasets.}
\vspace{-0.5em}

\begin{center}
\begin{tabular}{cccccc}
\toprule
Method & CIFAR100-50 & CIFAR100-80 & Stanford Cars & CUB & Aircraft \\ \midrule
Kmeans & 28.3${\pm 0.7}$ & 56.3${\pm 1.7}$ & 13.1${\pm 1.0}$ & 42.2${\pm 0.5}$ & 18.5${\pm 0.3}$ \\
DTC\cite{han2019learning} & 35.9${\pm 1.0}$ & 67.3${\pm 1.2}$ & - & - & - \\
RankStats+\cite{han2021autonovel} & 44.1${\pm 3.7}$ & 75.2${\pm 4.2}$ & 36.5${\pm 0.6}$ & 55.3${\pm 0.8}$ & 38.4${\pm 0.6}$ \\
NCL\cite{zhong2021neighborhood} & 52.7${\pm 1.2}$ & 86.6${\pm 0.4}$ & 43.5${\pm 1.2}$ & 48.1${\pm 0.9}$ & 43.0${\pm 0.5}$\\
ComEx\cite{yang2022divide} &53.4${\pm 0.7}$  & 85.7${\pm 1.3}$ & - & - & -\\
UNO\cite{fini2021unified} & 60.4${\pm 1.4}$ & 90.4${\pm 0.2}$ & 49.8${\pm 1.4}$ & 59.2${\pm 0.4}$ & 52.1${\pm 0.7}$ \\ 
GCD\cite{vaze2022generalized} & - & - & 42.6${\pm 0.4}$ & 56.4${\pm 0.3}$ & 49.5${\pm 1.0}$ \\ \midrule
Ours & \textbf{65.3}${\pm 0.6}$ & \textbf{91.2}${\pm 0.1}$ & \textbf{53.5}${\pm 0.8}$ & \textbf{65.7}${\pm 0.6}$ & \textbf{55.8}${\pm 0.9}$ \\ \bottomrule
\end{tabular}
\end{center}
\label{tab:train_novel}
\vspace{-1em}
\end{table*}

\subsection{Experimental Setup}
\paragraph*{Datasets:}
To evaluate the effectiveness of our method, we first conduct tests on the typical CIFAR100 dataset \cite{krizhevsky2009learning}. Specifically, we divide CIFAR100 into two categories: 80/20 known/novel classes and 50/50 known/novel classes, with the latter being more challenging. As the results on CIFAR10 \cite{krizhevsky2009learning} and ImageNet\cite{deng2009imagenet} datasets are nearly saturated \cite{fini2021unified}, we turn to evaluate our method on three fine-grained datasets - Stanford Cars \cite{krause20133d}, CUB \cite{wah2011caltech}, and FGVC-Aircraft \cite{maji2013fine} - which are more demanding for novel class discovery. We divide these datasets into two halves, with one comprising known classes and the other consisting of novel classes. The details of the dataset splits are presented in Tab.\ref{exp:dataset}.

\begin{table*}[!t]\small
\caption{Comparison with state-of-the-art methods on CIFAR100, Stanford Cars, CUB, and Aircraft datasets under the inductive setting, using task-agnostic evaluation protocol. GCD \cite{vaze2022generalized} is not applicable to the test set.}
\vspace{-0.5em}
\begin{center}
\begin{tabular}{c|ccc|ccc|ccc|ccc}
\toprule
 & \multicolumn{3}{c|}{CIFAR100-50} & \multicolumn{3}{c|}{Stanford Cars} & \multicolumn{3}{c|}{CUB} & \multicolumn{3}{c}{Aircraft} \\
Method & Known & Novel & All & Known & Novel & All & Known & Novel & All & Known & Novel & All \\ \midrule
RankStats+ \cite{han2019learning} & 69.7 & 40.9 & 55.3 & 81.8 & 31.7 & 56.3 & 80.7 & 51.8 & 66.1 & 66.4 & 36.5 & 51.5 \\
NCL \cite{zhong2021neighborhood}& 72.4 & 25.7 & 49.0 & 83.5 & 24.4 & 53.4 & 79.8 & 13.1 & 46.3 & 62.8 & 26.5 & 44.6 \\
UNO \cite{fini2021unified}& 75.0 & 57.6 & 66.3 & 81.7 & 46.7 & 63.9 & 78.7 & 62.1 & 70.3 & 71.2 & 52.4 & 61.8 \\
Ours & \textbf{78.6} & \textbf{59.4} & \textbf{69.0} & \textbf{83.9} & \textbf{51.3} & \textbf{67.3} & \textbf{81.1} & \textbf{67.5} & \textbf{74.2} &  \textbf{72.2} & \textbf{55.2} & \textbf{63.7} \\ \bottomrule
\end{tabular}
\end{center}
\label{tab:test}
\vspace{-1.5em}
\end{table*}
\vspace{-1.0em}
\paragraph*{Metric:}
Following \cite{fini2021unified}, we evaluate our method in the transductive learning setting and inductive learning setting. In the transductive learning setting, we employ ClusterAcc to evaluate the train novel datasets. The formula for ClusterAcc is as follows:
\vspace{-0.5em}
\begin{equation}
    \text{ClusterAcc} = \mathop{max}_{perm \in P} \frac{1}{N} \sum_{i=1}^{N} \mathbf{1} \{y_i= perm (\hat{y}_i)\}
\end{equation}
where $y_i$ and $\hat{y}_i$ denote the ground-truth and predicted labels, respectively, while $P$ represents the set of all permutations. We use the Hungarian algorithm to find the best permutation. In the inductive learning  setting, we use the task-agnostic evaluation protocol \cite{fini2021unified} to evaluate model performance. The accuracy of the labeled data is calculated by using the standard accuracy metric. For unlabeled data, we use ClusterAcc to measure the performance of class discovery. This protocol is applied to a test set that includes both known and novel class data. As this evaluation does not require any prior knowledge of whether the data is novel or known, it is more suitable for real-world applications than the first type of evaluation.
%, ensuring the known and novel classes are evaluated correctly.
\vspace{-1.0em}
\paragraph*{Implementation Details:}
For the CIFAR100 dataset, following previous work \cite{han2021autonovel, fini2021unified}, we use ResNet18 as our backbone network. First, we pretrain the network on the known classes for 100 epochs, and then we jointly train our network on the known and novel classes for 500 epochs, which is similar to UNO \cite{fini2021unified}. The optimizer is SGD, and the learning rate first grows linearly and then cosine decays. We implement NCL \cite{zhong2021neighborhood} on the CIFAR100-50 setting, and other results are mostly cited from their papers. For Stanford Cars, CUB, Aircraft datasets, we utilize DINO \cite{caron2021emerging} pretrained ViT as our backbone, and we only finetune the last block of ViT by Adamw optimizer. For a fair comparison, we adopt the same training setting to implement RankStats \cite{han2021autonovel}, NCL \cite{zhong2021neighborhood} and UNO \cite{fini2021unified} on those datasets based on their released code, and the training is all converged. 
% For all datasets, we first initialize the network on the known classes, and then train the network on the known and novel classes jointly. 
Specially, we first pretrain the network on the known classes for 50 epoch, and then jointly train the network on the known and novel classes for 100 epochs. Regarding hyperparameters, we adhere to the settings outlined in \cite{fini2021unified} and \cite{hinton2015distilling}, assigning a value of 0.1 to $\tau$, 1 to $\alpha$ and 0.4 to $t$ for all datasets. Additionally, we set $\beta$ to 0.1 for most datasets, except CIFAR100-80, which we set to 0.02.
% fine-grained datasets with strong class relationships, and 0.05 for generic datasets with relatively weaker class relationships.
Furthermore, we analyze its sensitivity in the experiments. More comprehensive details are in Appendix.

\subsection{Results}
\paragraph*{Comparison with SOTA:} We compare our method with currently state-of-the-art, including UNO \cite{fini2021unified}, ComEx \cite{yang2022divide}, NCL \cite{zhong2021neighborhood} and RankStats+ \cite{han2021autonovel} in Tab.\ref{tab:train_novel}. Following them, we first present the results on the typical CIFAR100 dataset. Our result on the CIFAR100-50 dataset significantly outperforms SOTA. In particular, we outperform UNO by 4.9\%. 
% Due to the results on CIFAR10 and ImageNet882-30 being nearly saturated, we introduce three fine-grained datasets, which is more challenging. And we reimplement UNO \cite{fini2021unified}, NCL~\cite{zhong2021neighborhood}, RankStats+ \cite{han2021autonovel}, GCD~\cite{vaze2022generalized} and Kmeans on those datasets for comparison. 
On three challenging fine-grained datasets, our experimental results demonstrate that our proposed approach outperforms UNO \cite{fini2021unified} by 3.7\%, 6.5\%, and 3.7\% on Stanford Cars, CUB, and Aircraft datasets, respectively. 

In addition, as shown in Tab.\ref{tab:test}, we report the results on the test dataset with task-agnostic evaluation metric. Overall, our method outperforms previous methods by a significant margin on both known and novel classes. Specifically, on the typical CIFAR100-50 dataset, we achieve 3.6\% and 1.8\% improvement on known and novel classes, respectively. On three challenging fine-grained datasets, our method outperforms the previous state-of-the-art (UNO) by 3-5\% on novel classes, while also obtaining 1-2\% gains on known classes. We hypothesize that our method enhances the learning of novel classes by leveraging class relationships, resulting in a direct improvement in the clustering effectiveness of novel classes, as well as a reduction in noise during the learning process. This, in turn, enhances representation learning and indirectly improves the performance of known classes.
In conclusion, the strong performance of our method on those challenging datasets demonstrates the effectiveness of our approach.

% \paragraph*{Results on Test set:}

% \begin{table*}[!t]\small
% \begin{center}
% \begin{tabular}{c|ccc|ccc|ccc|ccc}
% \toprule
%  & \multicolumn{3}{c|}{CIFAR100-50} & \multicolumn{3}{c|}{Stanford Cars} & \multicolumn{3}{c|}{CUB} & \multicolumn{3}{c}{Aircraft} \\
% Method & Known & Novel & All & Known & Novel & All & Known & Novel & All & Known & Novel & All \\ \midrule
% RankStats+ & 69.7 & 40.9 & 55.3 & \textbf{82.2} & 32.5 & 56.9 & \textbf{80.5} & 49.6 & 65.0 & 66.1 & 37.5 & 51.9 \\
% UNO & 75.0 & 57.6 & 66.3 & 81.6 & 45.8 & 63.4 & 78.7 & 63.1 & 70.9 & \textbf{70.8} & 52.4 & 61.6 \\
% Ours & \textbf{76.9} & \textbf{62.3} & \textbf{69.6} & 77.7 & \textbf{53.0} & \textbf{65.2} & 71.5 & \textbf{73.0} & \textbf{72.2} & 69.2 & \textbf{58.7} & \textbf{63.9} \\ \bottomrule
% \end{tabular}
% \end{center}
% \caption{Comparison with state-of-the-art methods on CIFAR-100, Stanford Cars, CUB, and Aircraft on both labeled and unlabeled classes, using task-agnostic evaluation protocol. Accuracy and clustering accuracy are reported on the test set.}
% \end{table*}

\begin{figure*}[!tbp]
\centering
\includegraphics[width=0.8\linewidth]{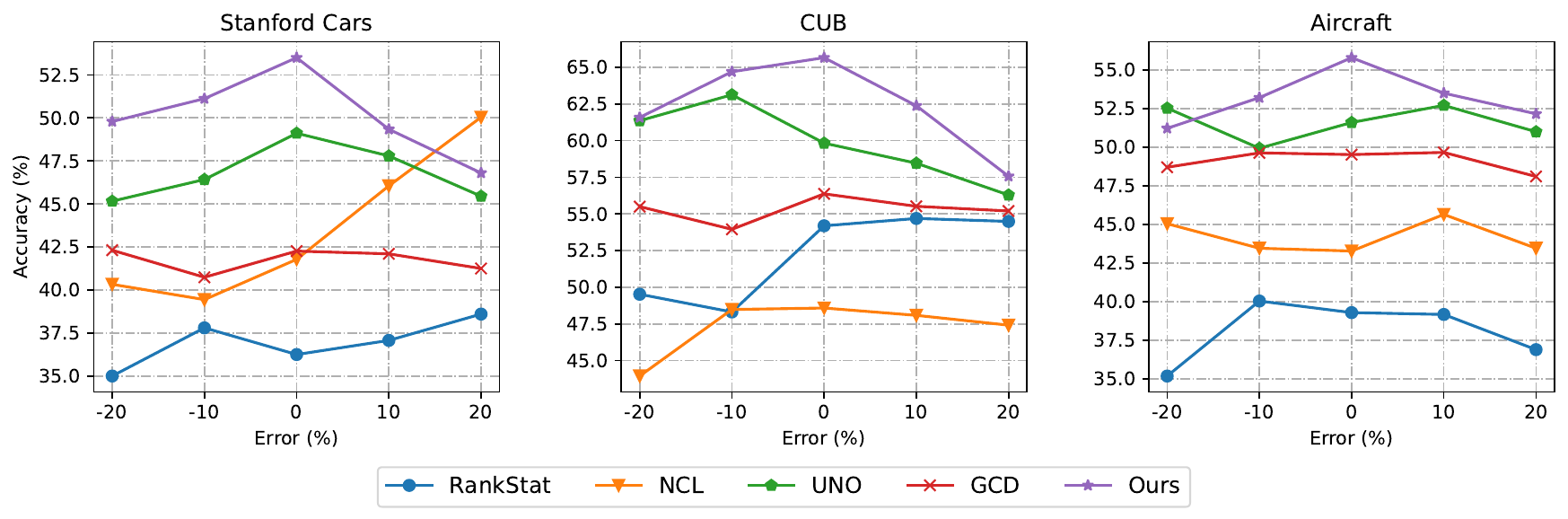}
\caption{The x-axis in these plots represents the error rate of the estimated number of novel clusters. For the CUB dataset, which has 100 novel classes, -20\% and 20\% denote underestimated and overestimated 20 classes, respectively.}
\label{fig:unknown_cluster}
\vspace{-1em}
\end{figure*}
\vspace{-1.0em}
\paragraph*{The number of clusters is unknown:} The above experiments assume that the number of novel classes is known. However, this is unrealistic in practice. Therefore, in order to further validate the effectiveness of our method in practical scenarios, we conduct experiments in the case of an unknown number of classes. We assume that the deviation between the estimated classes and the true classes is between -20\% and 20\%. For the case where there are 100 novel classes, -20\% means that the estimated classes are 20 less than the true classes, and +20\% means that the estimated classes are 20 more than the true classes. We report the results on the train novel dataset. As shown in Fig.\ref{fig:unknown_cluster}, in most cases, our method performs better than previous methods. Moreover, the more accurate the class estimation, the more obvious the advantage of our method. We speculate that when the class estimation error is relatively large, multiple novel classes may merge or split, making the relationship between the novel and known classes noisy and not conducive to learning relations.
% \begin{table}[!t]\small
% \begin{center}
% \begin{tabular}{c|ccc}
% \toprule
% Method & Stanford Cars & CUB & Aircraft \\ \midrule
% Kmeans & 11.9 & 34.2 & 17.2 \\
% RankStats+ & 35.0 & 43.6 & 31.9 \\
% NCL & 40.3 & 50.0 & 43.5 \\
% UNO & 45.2 & 51.3 & 48.0 \\ \midrule
% % GCD & 42.8 & 54.4 & 50.9 \\ \midrule
% Ours & \textbf{51.0} & \textbf{55.0} & \textbf{52.5} \\ \bottomrule
% \end{tabular}
% \end{center}
% \caption{Comparison with state-of-the-art methods on Stanford Cars, CUB, and Aircraft for novel class discovery with an unknown number of classes $C^u$, using task-aware evaluation protocol. Clustering accuracy is reported on the unlabeled train set.}
% \end{table}
\begin{figure}[!tbp]
\centering
\begin{minipage}{0.32\linewidth}
  \vspace{3pt}
  \centerline{\includegraphics[width=\textwidth]{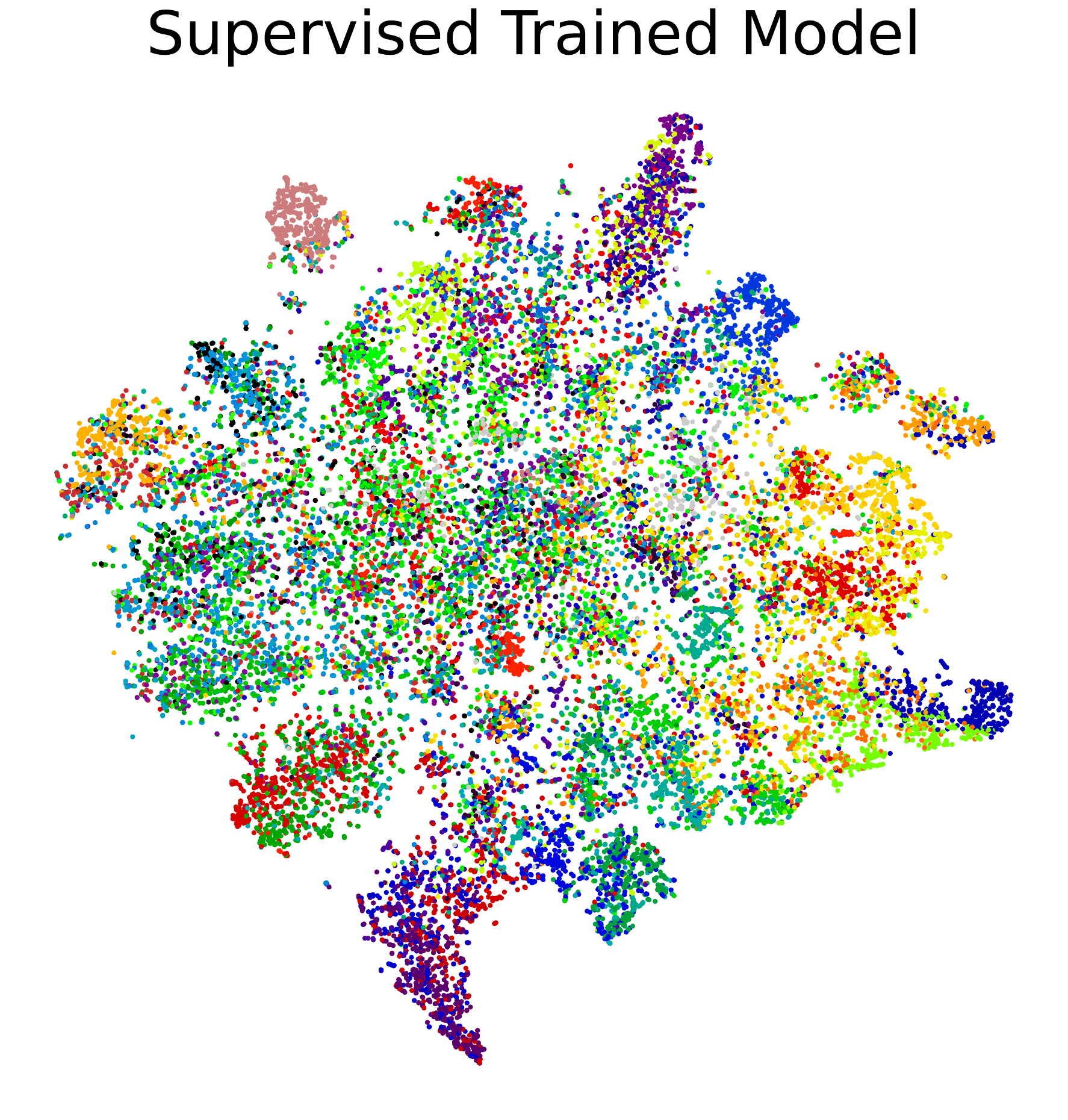}}
\end{minipage}
\begin{minipage}{0.32\linewidth}
\vspace{3pt}
\centerline{\includegraphics[width=\textwidth]{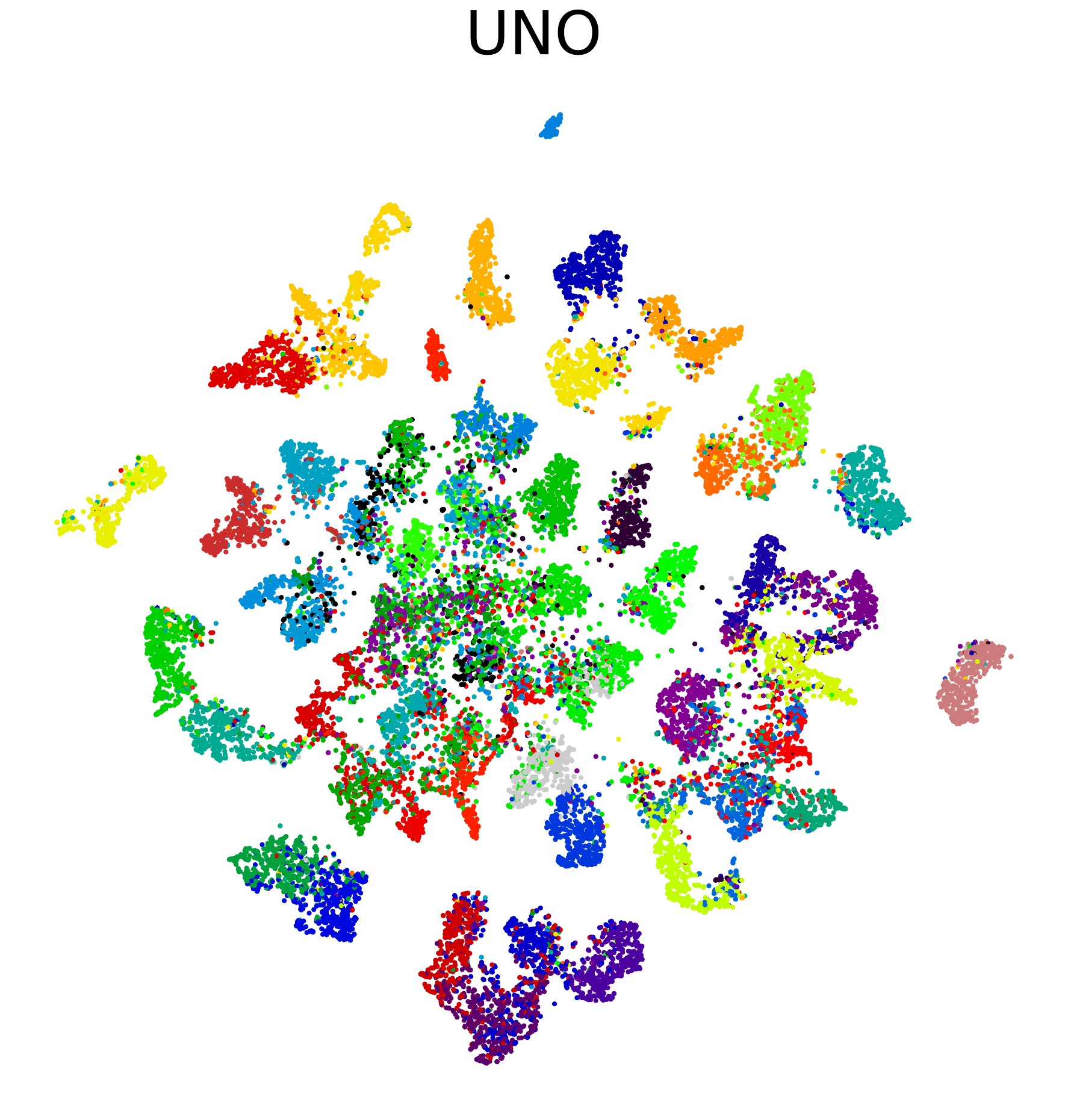}}
\end{minipage}
\begin{minipage}{0.32\linewidth}
\vspace{3pt}
\centerline{\includegraphics[width=\textwidth]{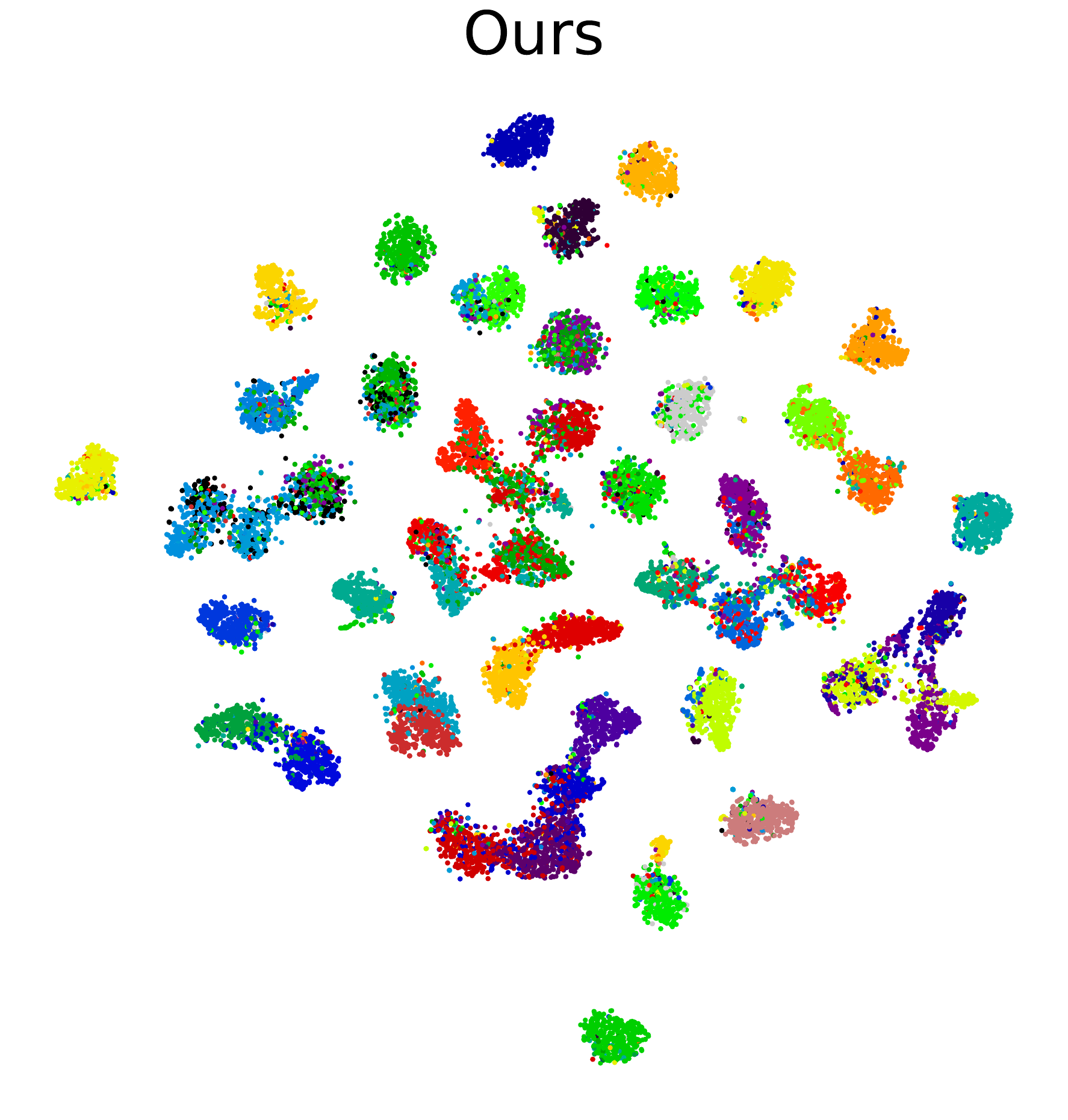}}
\end{minipage}
\caption{t-SNE visualization of unlabeled training set on CIFAR100-50. }\label{fig:t-SNE}
\vspace{-1em}
\end{figure}

\begin{figure}[!tbp]
\centering
\includegraphics[width=1.0\linewidth]{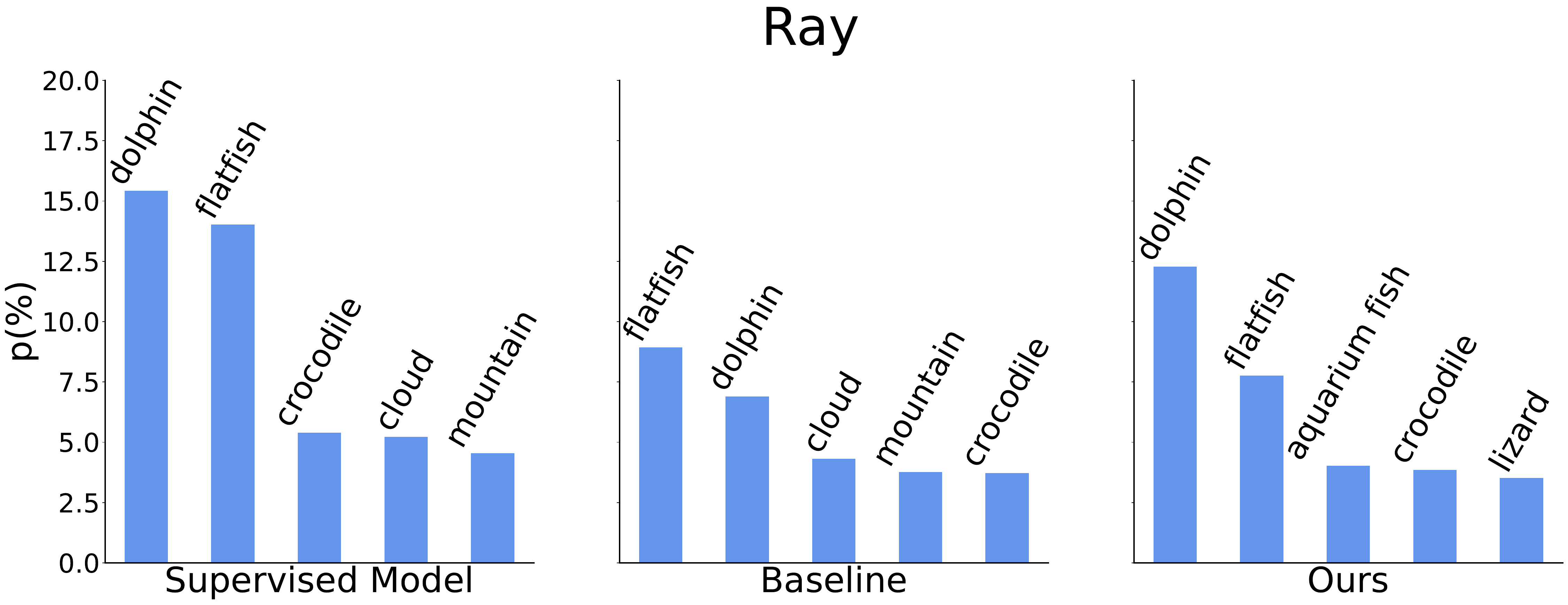}
\caption{Visualization of quantified relative relationships. The bar labels represent the known classes in the CIFAR100-50 setting. Each plot shows average predictions for instances of the novel ``ray" class on the known class head.}
\label{fig:relation}
\vspace{-1em}
\end{figure}

\vspace{-1.0em}
\paragraph*{Visualization:}
We conducted a qualitative analysis of the learned feature space using t-SNE \cite{JMLR:v9:vandermaaten08a}. As depicted in Fig.\ref{fig:t-SNE}, the supervised pre-trained model is noisy and some classes produced by UNO \cite{fini2021unified} are entangled, making it difficult for a linear classifier to distinguish the samples. In contrast, our proposed method generates more compact feature representations that tightly group samples of the same class. 
% Consistent with our quantitative findings, our approach outperforms the comparison method in separate classes.

% We report a qualitative analysis showing the feature space learned by our model on CIFAR-100-50. In Fig.\ref{fig:t-SNE}, we visualize the feature space of novel classes generated by the encoder and plot it using t-SNE\cite{JMLR:v9:vandermaaten08a} in two dimensions. We also apply the same procedure to the features produced by UNO\cite{fini2021unified} for comparison. From the plot, it is clear that our model produces feature representations for samples of the same class that are more tightly grouped compared to UNO\cite{fini2021unified}. At the same time, some classes in UNO\cite{fini2021unified} are entangled together, making it hard for a linear classifier to discriminate the samples. However, we note that this phenomenon occurs infrequently in our approach. In accordance with our quantitative results, our method does a better job of separating the classes.

We also present the class relationships produced by our model on CIFAR100-50. Fig.\ref{fig:relation} illustrates the averaged predictions for instances of the novel ``ray" class on the known class head. It shows that the baseline method fails to capture the ``dolphin \textgreater \ flatfish" relation order of ``ray", and while our proposed method preserves this order. Additionally, our approach exhibits superior performance in filtering out background noise, such as ``cloud" and ``mountain", and discovering innovative relationships between categories, such as ``aquarium fish" and ``lizard". More visualizations are available in the appendix.

% not only inherit the knowledge of supervised trained model

% but also effectively reduce the noise caused by the learning process of unsupervised classification.  

Overall, our findings demonstrate the effectiveness of our approach in learning feature representations and class relationships.

\begin{table}[!t]\small
\caption{Ablation study. $\mathcal{L}^u_{rKD}$ stands for adding KD loss on the unlabeled data, and $g(\eta)$ is the learnable weight function on $\mathcal{L}^u_{rKD}$. All results are evaluated on the unlabeled training set.}
\begin{center}
\begin{tabular}{cc|ccc}
\toprule
$\mathcal{L}^u_{rKD}$  & $g(\eta)$ & Stanford Cars & CUB & Aircraft\\
 \midrule
\usym{2717}  & \usym{2717} & 49.1 & 59.2 & 52.1  \\
\usym{2713} & \usym{2717} &51.9              &   63.7  &     53.4 \\
\usym{2713} & \usym{2713} & \textbf{53.5}             &   \textbf{65.7}  &     \textbf{55.8} \\
\bottomrule
\end{tabular}
\end{center}
\label{tab:ablation_1}
\vspace{-1em}
\end{table}

\begin{table}[!t]\small
\caption{Ablation the design of our learnable weight function $g(\eta)$. SG and Norm denote Stop Gradient and Normalization, respectively. All results are evaluated on the unlabeled training set.}
\begin{center}
\begin{tabular}{c|ccc}
\toprule
 $g(\eta)$ & Stanford Cars & CUB & Aircraft \\ \midrule
  1 & 51.9              &   63.7  &     53.4     \\
  $\eta$ & 32.1              &   58.5  &     \textbf{58.5}     \\
 $\text{SG}(\eta)$ &   51.2            &  64.3   &    54.7      \\
 $\text{SG}(\text{Norm}(\eta))$ &         51.9      &  64.6   &  54.2        \\
 $\text{Norm}(\eta)$ & \textbf{53.5}             &   \textbf{65.7}  &     55.8    \\ \bottomrule
\end{tabular}
\end{center}
\label{tab:ablation_2}
\vspace{-1em}
\end{table}
\subsection{Ablation study}
\paragraph*{Component Analysis:} 
As shown in Tab. \ref{tab:ablation_1}, we conduct an ablation study to evaluate the effectiveness of our novel class relation knowledge distillation regularization ($\mathcal{L}^u_{rKD}$) and learnable weight function ($g(\eta)$). With $\mathcal{L}^u_{rKD}$, we see improvements of 2.8\%, 4.5\%, and 1.3\% on the Stanford Cars, CUB, and Aircraft datasets, respectively, demonstrating that the class relation helps guide the learning of novel classes. Furthermore, the incorporation of the learnable weight function leads to even further improvement in all datasets, confirming the effectiveness of both modules.
\vspace{-0.5em}
\paragraph*{Learnable weight function:} 
% 我们的动机是对不同语义关系的样本，施加不同强度的约束。具体而言，对于那些语义关系比较强的样本，我们施加更多约束。因此，我们的权重是语义关系系数的正相关函数。本文中，我们尝试了几种简单的设计。如图所示，我们从stop gradient 和 normalization的角度，分析了几个不同的设计。例如SG(\eta)表示，我们采用g(\eta)=\eta，并且不回传梯度，对于g(\eta)=\eta，回传梯度存在退化解。比较SG(\eta)和1，我们发现SG(\eta)没有明显的提升，我们认为，在fine-grained的数据集上，新类与已知类之间的关系都比较强，不同样本的\eta差别很小，因此它和1的效果差不多。而SG(Norm(\eta))相较于1，还有些下降，我们认为，L \eta都 比较大的样本大概率是noise， normalize之后，样本之间的差距拉大了，这可能导致了一些noise 样本的作用被放大了，从而干扰了学习。而我们的方法，可以拉远L比较大的样本，使得\eta降低，从而降低了noise学习，此外，他可以拉进L比较小样本，使得模型充分学习共享的语义信息。
Our goal is to apply constraints with different intensities to samples that have varying semantic relationships. Specifically, we apply stronger constraints to samples with higher semantic relationships. 
%Therefore, the weight we assign is a positive correlation function of the semantic relationship coefficient.

In Table \ref{tab:ablation_2}, we investigate several simple designs for the weight function and analyze them from the perspectives of stop gradient ($\text{SG}$) and normalization ($\text{Norm}$). For instance, $\text{SG}(\eta)$ involves using $g(\eta)=\eta$ without backpropagating gradients. 
% We compare the $\text{SG}({\eta}), \text{SG}(\text{Norm}(\eta))$ method with naive 1 and observe improved results on the CUB and Aircraft datasets, validating the effectiveness of the weight function. 
We compare $\text{Norm}(\eta)$ with $\text{SG}(\text{Norm}(\eta))$, yielding significant improvement on all three datasets, indicating the superiority of the learnable characteristics. Moreover, $\text{Norm}(\eta)$ is more stable than $\eta$, and outperforms $\eta$ on Stanford Cars and CUB datasets, validating the effectiveness of normalization. 
Although there is theoretically a degenerate solution in our weight function, which assigns the maximum weight to the sample with the smallest KL loss term. In practice, due to the randomness of batch samples, the model is difficult to optimize towards this degenerate solution. More analysis is in the Appendix.

% this degenerate solution did not occur. We believe that it is difficult for the model to optimize to this degenerate solution. More analysis is in the Appendix.

% . This may be due to the fact that on fine-grained datasets, new and known class relationships are relatively strong, and the differences in $\eta$ between different samples are small, even after normalization. Therefore, its effect is similar to that of 1.
% % how to prove that?
% In comparison, $\text{SG}(\text{Norm}(\eta))$ decreases slightly in comparison to 1. We speculate that samples with larger values of $\mathcal{L}^u_{rKD}$ and $\eta$ are more likely to be noise. After normalization, the impact of the noise can be magnified, which might impede the learning process. 
% To address this possible issue, we propose a learnable weight function to encourage the diversity of $\eta$ for different samples. In particular, we pull away samples with relatively large $\mathcal{L}^u_{rKD}$ by reducing $\eta$. Additionally, our method can pull in samples with relatively small $\mathcal{L}^u_{rKD}$, which enhances the learning of shared semantic information more effectively by the model. 
% Additional theoretical analysis is provided in the Appendix.

\vspace{-0.5em}
\paragraph*{Hyperparameter $\beta$:}
% 我们的方法简单有效，它只引入了一个额外的超参$\beta$，它用来控制class relation regularization term的强弱。如Tab.6所示，在Train novel，相较于beta=0，不同的beta值都有明显的提升，而且beta越大，提升越明显。但是beta太大，会使模型过度专注于novel class的学习，使得known class的学习比较弱。在测试集上，降低了模型在known上的性能，使得整体性能下降。因此，我们推荐使用0.1作为默认参数，平衡known class和novel class的学习。
Our proposed approach is simple and effective, utilizing a single hyperparameter $\beta$ to regulate the impact of the class relation regularization term. As shown in Tab.\ref{tab:ablation_3}, our experiments on the unlabeled training dataset (Train-Novel) demonstrate that various values of $\beta$ result in significant improvements over the baseline model with $\beta=0$. Furthermore, we observe that as the value of $\beta$ increases, the relative gains become increasingly conspicuous. However, it is important to note that excessively high values of $\beta$ may produce a model that overemphasizes novel class learning at the expense of weaker performance in the known classes. This, in turn, can lead to lower overall performance during testing. Based on our results, we suggest using a default value of $\beta=0.1$, which achieves a better balance between known and novel class learning.
\begin{table}[!t]\small
\caption{Analysis of hyperparameter $\beta$. ``Train-novel" refers to the evaluation of an unlabeled training dataset, whereas the other results are evaluations of the test set.}
\begin{center}
\resizebox{0.35\textwidth}{!}{
\begin{tabular}{c|cccc}
\toprule
\multirow{2}{*}{$\beta$} & \multicolumn{4}{c}{CUB} \\
&  Train-Novel & Known & Novel & All \\ \midrule
  0 & 59.2              &   78.7  &     62.1   & 70.3  \\
  0.01 & 62.1              &   80.5  &     63.1   & 71.8  \\
  0.02 & 62.0              &   80.4  &     64.0   & 72.2  \\
  0.05 & 63.6              &   81.0  &     66.5   & 73.7  \\
    0.1 & 65.7              &   \textbf{81.0}  &     67.5   & \textbf{74.2}  \\
  0.2 & 65.1             &   80.3  &     65.8   & 73.0  \\
  0.5 & 67.3              &   78.3  &     67.4   & 72.8  \\ 
1 & \textbf{67.9}              &   77.2  &     \textbf{68.4}   & 72.8  \\
\bottomrule

\end{tabular}}
\end{center}
\label{tab:ablation_3}
\vspace{-2em}
\end{table}

\section{Conclusion}
% 本文中，我们提出了一个novel class-relation knowledge distillation learning framework, which provides a new perspective to transfer knowledge from known to novel clases for the NCD community. Instead of transferring knowledge only by sharing representation space, we utilize class relations to transfer knowledge. In particular, 我们观察到novel classes在known classes训练地模型上响应，能够很好的表示novel classes与known classes之间的关系，但是，这种关系在discovery traing phase被破坏了。因此，为了保持这种meaningful的类间关系，我们提出了一个简单有效的regularization term，它约束discovery training phase的模型，保持原来的关系。并且，我们提出一个可学习的权重函数，它动态地分配更多的权重给语义相近的样本，使模型学习共享的语义信息。我们在几个通用数据集，和几个fine-grained dataset上都取得了明显的提升，验证了我们方法的有效性。此外，我们希望我们的发现，能够shed more light on futre work to explore the relation between known and novel classes, and enhance the models transferbility.

In this paper, we propose a novel class-relation knowledge distillation learning framework, which provides a new perspective to transferring knowledge from known to novel classes in the NCD problem. Instead of transferring knowledge only by sharing representation space, we utilize class relations to transfer knowledge. Specifically, we observe that the prediction distribution of novel classes on a model trained on known classes effectively captures the relationship between the novel and known classes. However, this relationship is disrupted during the discovery training stage. Therefore, to maintain this meaningful inter-class relationship, we propose a simple and effective regularization term that constrains the model in the discovery training stage. Additionally, we propose a learnable weight function that dynamically assigns more weight to semantically similar samples, enabling the model to learn the shared semantic information. Our method achieves significant improvements on several general datasets and fine-grained datasets, validating the effectiveness of our approach. Furthermore, we hope our findings will shed more light on future work to explore the relationship between known and novel classes and enhance the model's transferbility for knowledge transfer.

\section*{Acknowledgement}
This work was supported by Shanghai Science and Technology Program 21010502700, Shanghai Frontiers Science Center of Human-centered Artificial Intelligence and MoE Key Lab of Intelligent Perception and Human-Machine Collaboration (ShanghaiTech University).
{\small
\bibliographystyle{ieee_fullname}
\bibliography{egbib}
}
\clearpage
\appendix
\section{Implementation Details}
For the CIFAR100 dataset, we adopt the SGD optimizer and employ a learning rate schedule that initially increases from 0.001 to 0.4 within the first 10 epochs and then decreases to 0.001 at 500 epochs using a cosine annealing schedule. Our batch size is 512. We are re-implementing NCL \cite{zhong2021neighborhood} on the CIFAR100-50 setting using their publicly available code, and we cite all other results from their published work.

As for the three fine-grained datasets, we use the Adamw optimizer, and our learning rate scheduling involves an initial increase from 0.0001 to 0.001 within ten epochs, followed by a decrease to 0.0001 at 100 epochs using a cosine annealing schedule. We utilize a batch size of 512 for all methods and reimplement the results of GCD \cite{vaze2022generalized} using the code they provided.

To enhance the performance of our clustering approach and for a fair comparison, we also employed a multi-head technique similar to UNO. We used four heads for the CIFAR100 dataset and two heads for the remaining three fine-grained datasets. Our novel class head includes a Multilayer Perceptron (MLP) and a cosine classifier.

Moreover, we determine the hyperparameter $\beta$ through the validation set on the known class.

\section{More visualization}

To demonstrate the efficacy of our model, we selected the five representative novel classes and analyzed their relationships with known classes. The top four classes were selected based on their similarity to known classes, ordered from high to low, while the last class ``wardrobe," was selected from special novel classes that will often appear in images together with some known classes. As shown in Fig.\ref{fig:relations}, in many cases, the distributions generated by the supervised trained model have strong semantic information. Furthermore, the plot indicates that our model can better maintain relations between the novel and known classes than the baseline model(UNO). What's more, we analyze on the fine-grained datasets. As shown in figure, our model can inherit the relationship between the ``Spitfire'' class and known classes captured by the supervised trained model, while the baseline model loses this ability. Specifically, our model can capture the common characteristics of the Spitfire, C-47, C-130, and Cessna 208, such as propellers and forward wings. It also recognizes the unique color present in both the Spitfire and C-47. In contrast, the baseline model basically regards all known classes as the same. This shows that our model can well capture the potential relationship between novel classes and known classes on fine-grained datasets.

\begin{figure*}[!tbp]
    \centering
    \includegraphics[width=1.0\linewidth]{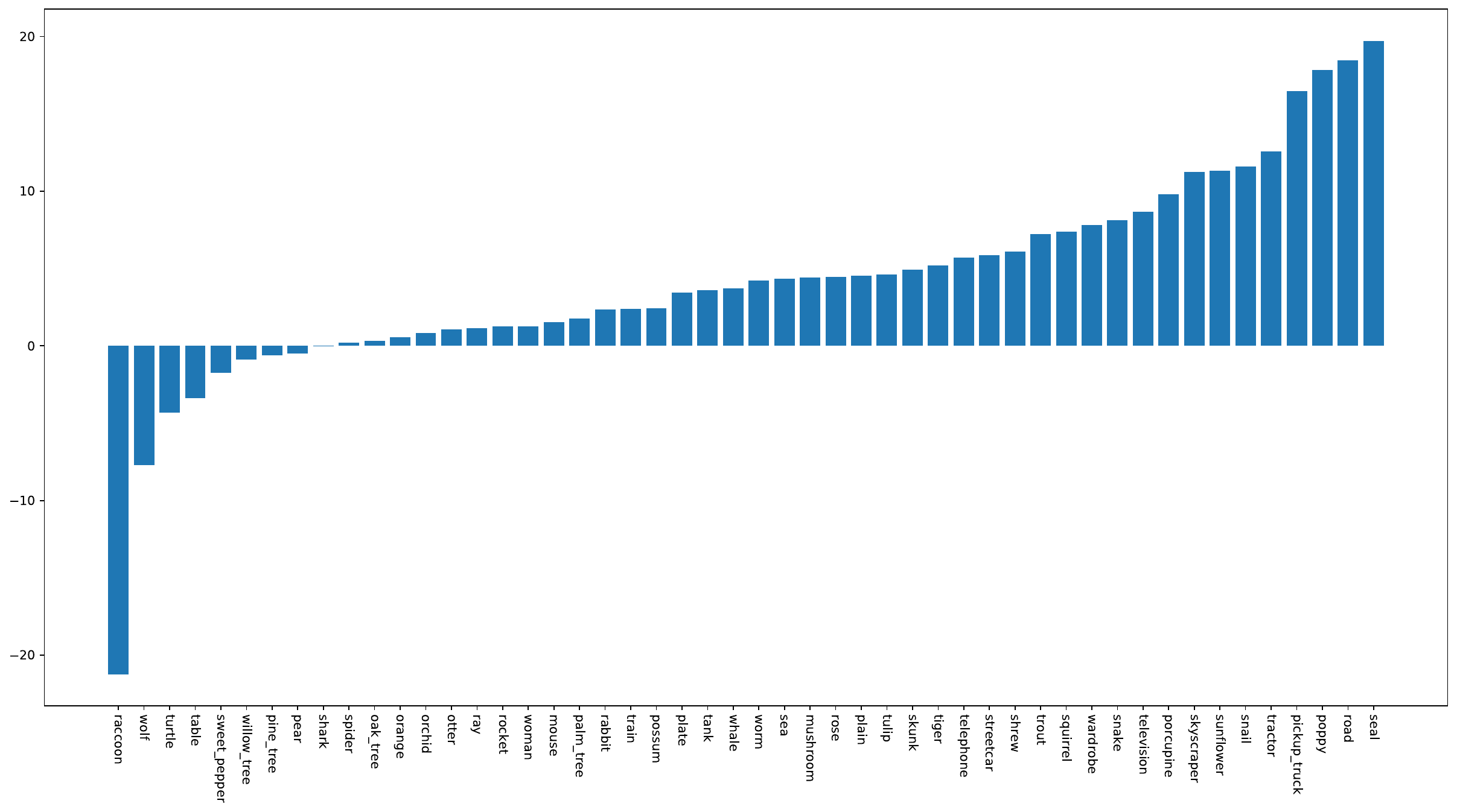}
    \caption{The difference between the accuracy of our model and the UNO model on each novel class.}
    \label{fig:acc}
\end{figure*}

In addition, we present a comparative analysis of the accuracy of our model and the baseline model for each novel class. Fig.\ref{fig:acc} demonstrates that our model's predictions are more accurate than the baseline model in almost all classes.

\begin{figure}[!tbp]
\centering
\subfigure{
\includegraphics[width=0.5\textwidth]{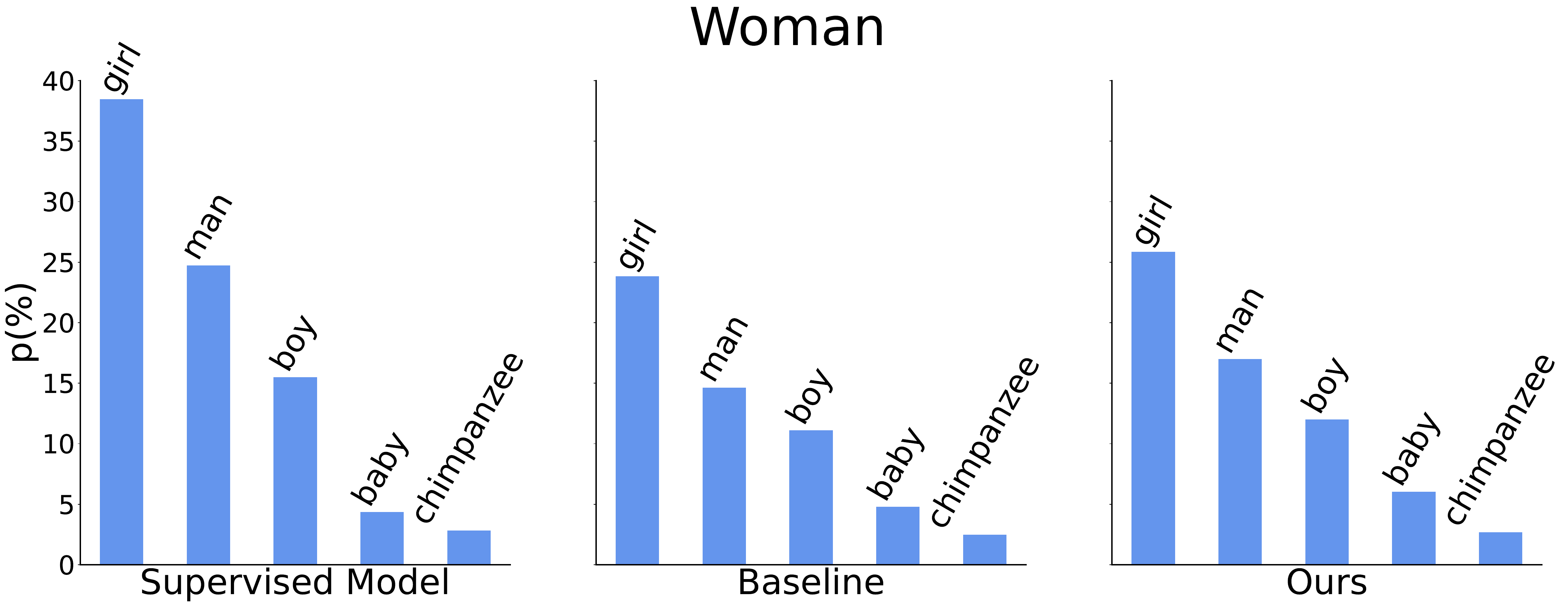}
}
\quad
\subfigure{
\includegraphics[width=0.5\textwidth]{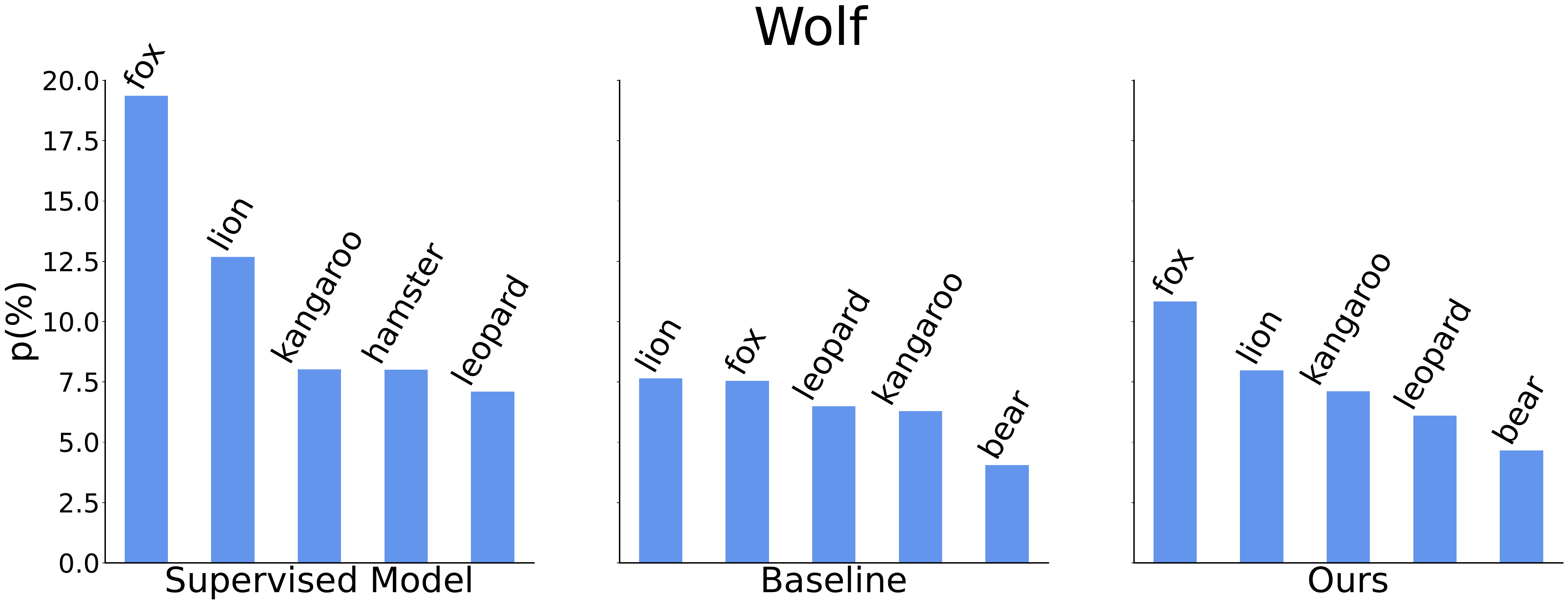}
}
\quad
\subfigure{
\includegraphics[width=0.5\textwidth]{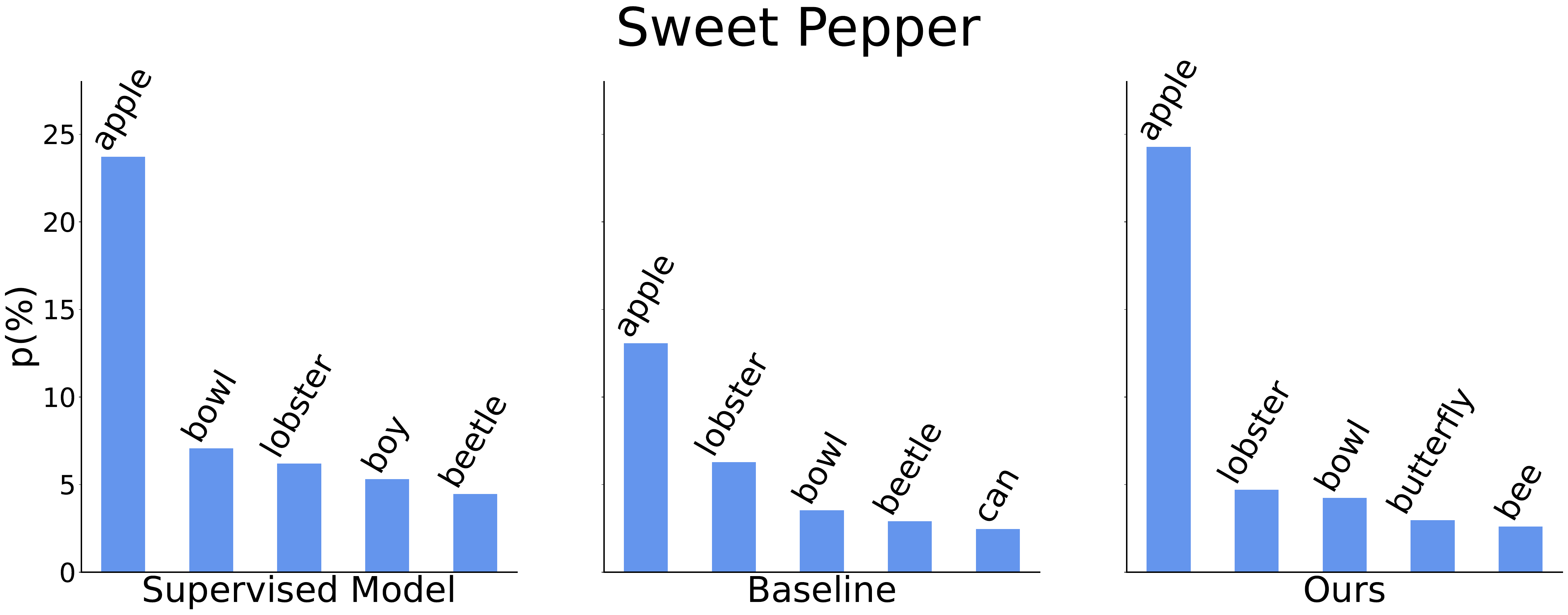}
}
\quad
\subfigure{
\includegraphics[width=0.5\textwidth]{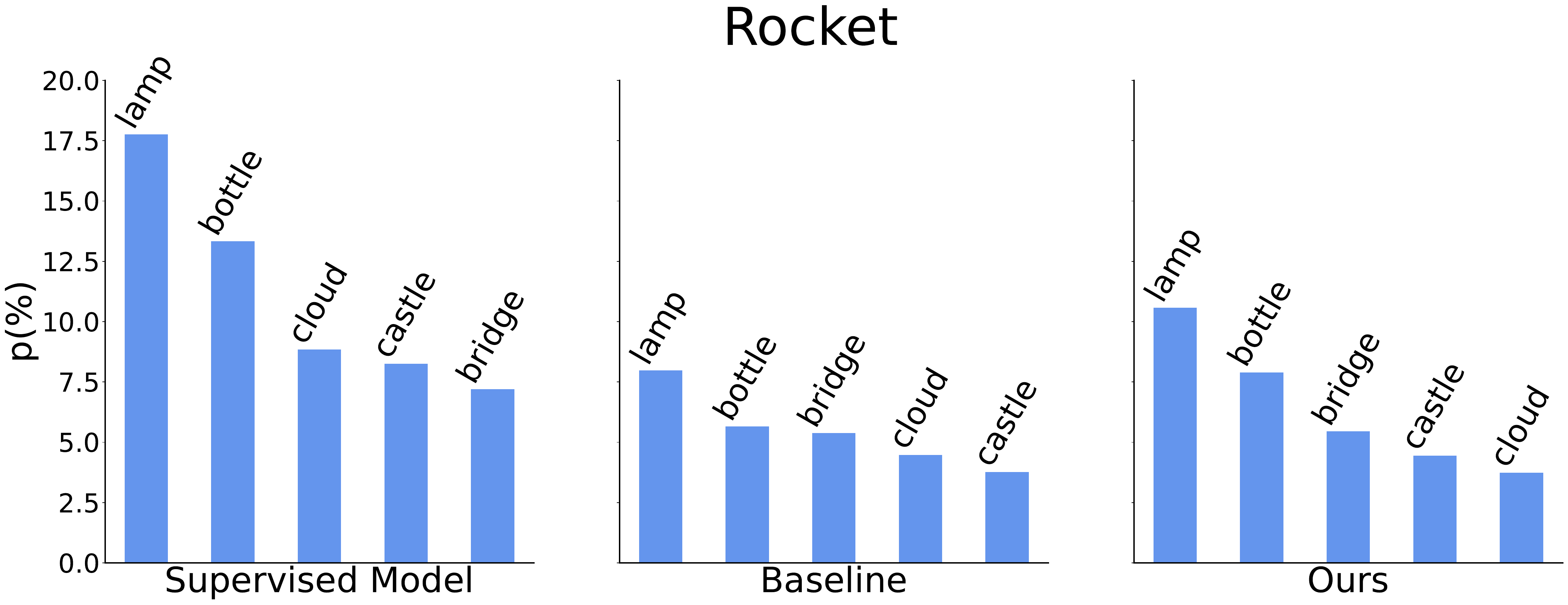}
}
\quad
\subfigure{
\includegraphics[width=0.5\textwidth]{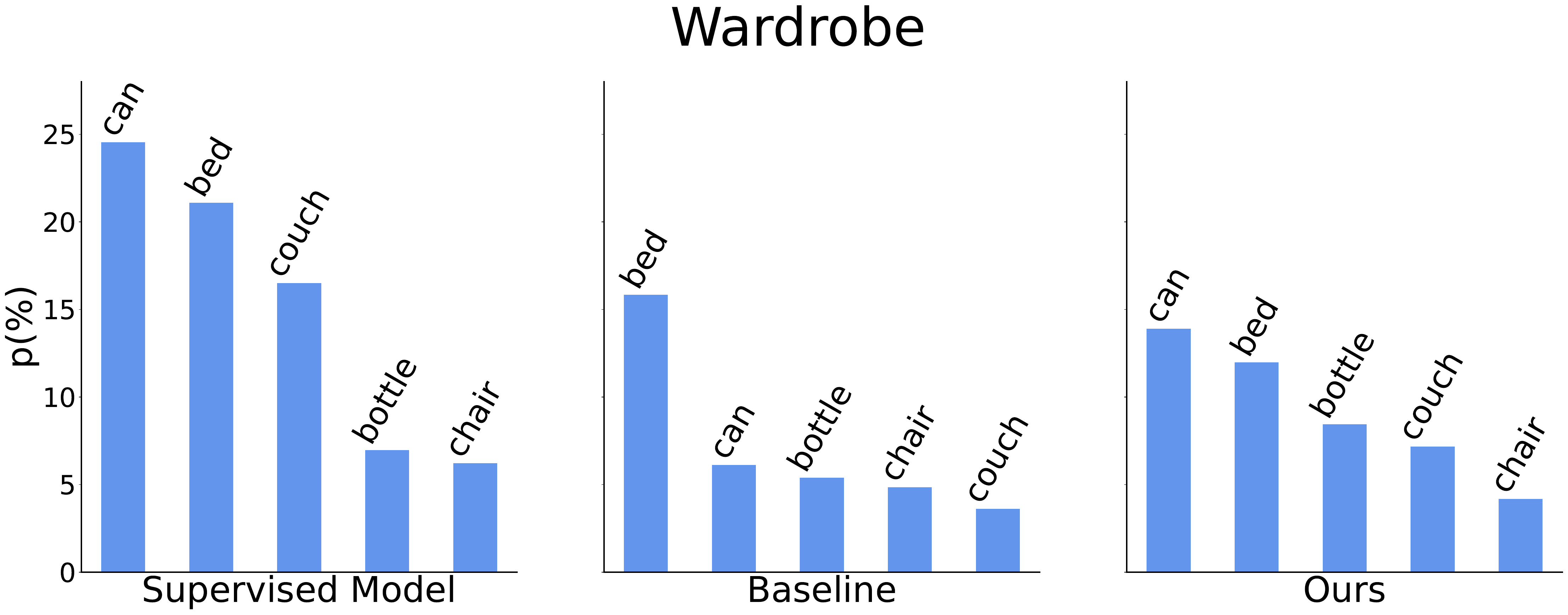}
}
\caption{Visualization of quantified relative relationships. The bar labels represent the known classes in the CIFAR100-50 setting. Each plot shows average predictions for instances of a novel class on the known class head. In most cases, our model's predictions are more similar to the supervised trained model's predictions than the baseline model's predictions.}\label{fig:relations}
\vspace{-1em}
\label{fig:relations}
\end{figure}
\begin{figure}[!h]
    \centering
    \includegraphics[width=1.0\columnwidth]{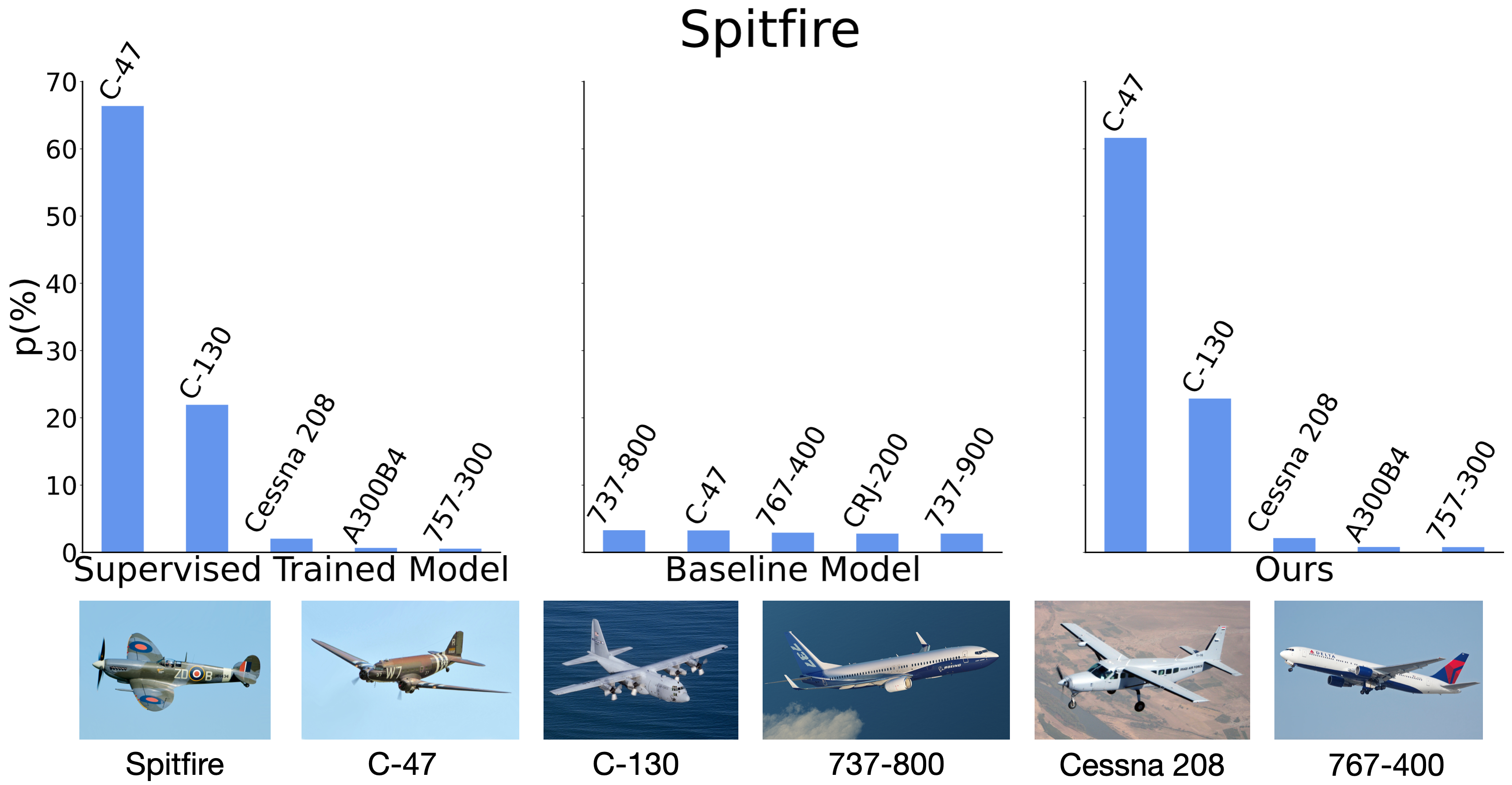}
    \caption{Visualization of relative relationships on Aircraft.}
    \label{fig:vis_aircraft}
\end{figure}

\section{More experiments without pre-trained model}
We conduct experiments on fine-grained datasets with ResNet18 from scratch. As the Tab.\ref{tab:train_novel} shows, we still achieve sizeable improvement over existing methods, 2.1\% on Stanford Cars, 1.8\% on CUB, and 6.0\% on Aircraft. This demonstrates that our method is also effective without pre-trained models.
\begin{table}[!h]\small
    \caption{Pre-train ResNet18 on known classes, and then train on known+novel classes. Both stages are trained for 200 epochs.}
    \begin{center}
    \resizebox{0.35\textwidth}{!}{
    \begin{tabular}{cccc}
    \toprule
    Method  & Stanford Cars & CUB & Aircraft \\ \midrule
    RankStats+ & 17.7 & 20.2 & 30.1 \\
    NCL & 27.1 & 24.7  & 36.7\\
    UNO & 25.2 & 24.5 & 37.8 \\ 
    Ours & \textbf{29.2} & \textbf{26.5} & \textbf{43.8} \\ \bottomrule
    \end{tabular}}
    \end{center}
    \label{tab:train_novel}
    \end{table}

\section{Temperature hyperparameter $T$ analysis}
We conducted an in-depth analysis of the temperature $T$ as presented in S\ref{tab:temp}. The results indicate that temperatures 1, 2, 4, and 6 yield satisfactory performance, demonstrating the robustness of our model with respect to the temperature hyperparameter. Taking into consideration the established practices in knowledge distillation \cite{hinton2015distilling,tian2019contrastive}, where the temperature is often set to 4, we have chosen this value for our model.
\begin{table}[!t]\small
    \caption{Results on CIFAR100-50 with different temperature.}
    % \vspace{-1em}
    \begin{center}
    \resizebox{0.4\textwidth}{!}{
    \begin{tabular}{c|ccccc}
    \toprule
    Temperature  & 1 & 2 & 4 & 6 & 8 \\ \midrule
    Novel Acc & 65.4 & 66.8 & 65.3 & 66.2 & 61.6\\ \bottomrule
    \end{tabular}}
    \end{center}
    \label{tab:temp}
    \end{table}

\section{Learnable weight function}
\begin{figure}[!tbp]
    \centering
    \includegraphics[width=1.0\linewidth]{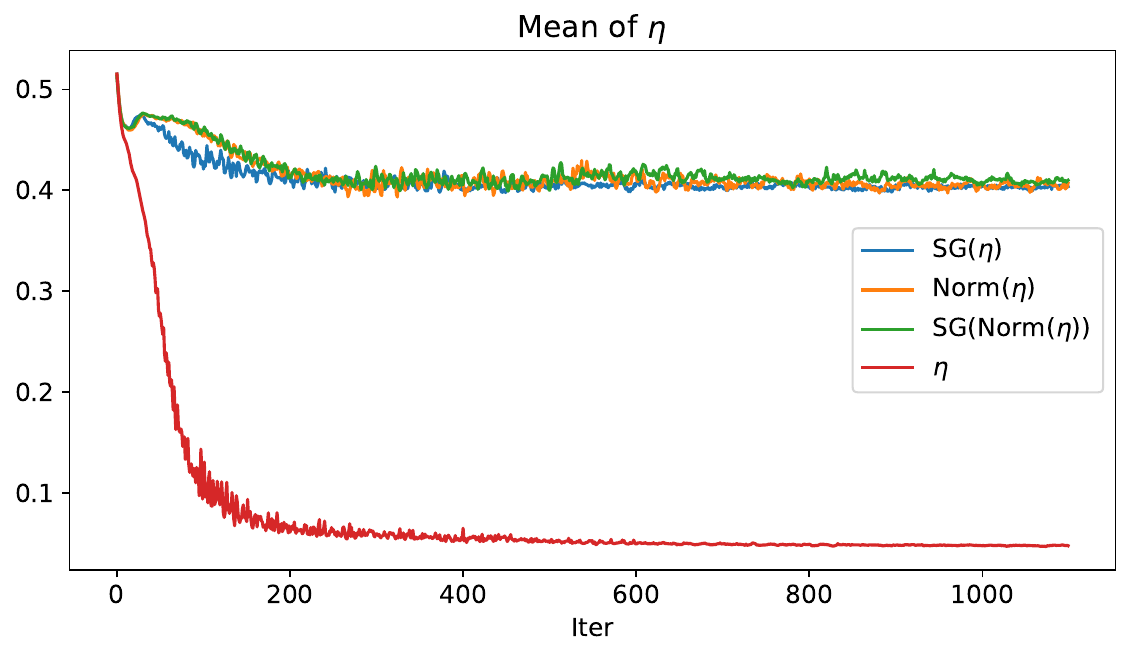}
    \caption{The mean of $\eta$ for different weight function $g(\eta)$.}
    \label{fig:eta}
\end{figure}

In the ablation study, we analyze various designs of our learnable weight function and demonstrate the superiority of our approach. However, the theoretically learnable weight function may have a degenerate solution, where the maximum weight is assigned to the sample with the smallest KL. Achieving this degenerate solution in practice is challenging due to the random selection of samples in each batch. Additionally, as shown in Fig.\ref{fig:eta}, the mean statistics of eta remain relatively stable when the batch size is large.

\section{Discussion with NCDwF \cite{joseph2022novel}}
In NCDwF, they focus on novel class discovery without forgetting, where known class data is not available in the discovery stage. To transfer knowledge, they introduce a mutual information regularization term for novel classes to couple the learning of labeled head to unlabeled head and expect to transfer semantic knowledge from known classes to novel classes. Meanwhile, known and novel classes still share a feature extractor. Differently, we transfer knowledge from a known classes pretrained model to a discovery trained model and expect the discovery trained model to maintain meaningful class relations. What's more, we also develop a simple and effective learnable weight function, which adaptively promotes knowledge transfer based on the semantic similarity between the novel and known classes. In addition, the outstanding results on a challenging dataset in Tab.\ref{tab:mi} show the superiority of our method. In conclusion, our method is totally different from NCDwF.

\begin{table}[]
    \begin{center}
        \begin{tabular}{c|cc}
            \toprule
            Method & CIFAR100-80 &  CIFAR100-50 \\ \midrule
            UNO    & 90.4 & 60.4       \\ 
            NCDwF  & 91.3 & 61.2       \\ 
            Ours   & 91.2 & 65.3       \\ \bottomrule
            \end{tabular}
    \end{center}
    \caption{The results on CIFAR100 dataset under the same setting.}
    \label{tab:mi}
\end{table}

\end{document}